\documentclass{article}
\usepackage{authblk}

\usepackage{graphicx}%
\usepackage{multirow}%
\usepackage{amsmath,amssymb}
\usepackage{amsthm}%
\usepackage{mathrsfs}%
\usepackage{url}

\usepackage{algorithmicx}%
\usepackage[ruled,vlined]{algorithm2e}

\usepackage{nameref}
\usepackage{subcaption}

\title{Clinical-ComBAT: a diffusion-weighted MRI harmonization method for clinical applications} 

\author[1,2]{Gabriel Girard}
\author[1,2]{Manon Edde}
\author[1,2]{Félix Dumais}
\author[1]{Yoan David}
\author[3]{Matthieu Dumont}
\author[3]{Guillaume Theaud}
\author[3]{Jean-Christophe Houde}
\author[2]{Arnaud Boré}
\author[2,3]{Maxime Descoteaux}
\author[1,2]{Pierre-Marc Jodoin}

\affil[1]{\small Videos \& Images Theory and Analytics Lab (VITAL), Department of Computer Science, Université de Sherbrooke, Sherbrooke (Qc), Canada}
\affil[2]{Sherbrooke Connectivity Imaging Lab (SCIL), Department of Computer Science, Université de Sherbrooke Sherbrooke (Qc), Canada}
\affil[3]{Imeka Solutions inc, Sherbrooke (Qc), Canada}

\begin{document}
\maketitle

\begin{abstract}
Diffusion-weighted magnetic resonance imaging (DW-MRI) derived scalar maps are effective for assessing neurodegenerative diseases and microstructural properties of white matter in large number of brain conditions. However, DW-MRI inherently limits the combination of data from multiple acquisition sites without harmonization to mitigate scanner-specific biases. While the widely used ComBAT method reduces site effects in research, its reliance on linear covariate relationships, homogeneous populations, fixed site numbers, and well populated sites constrains its clinical use. To overcome these limitations, we propose Clinical-ComBAT, a method designed for real-world clinical scenarios. Clinical-ComBAT harmonizes each site independently, enabling flexibility as new data and clinics are introduced. It incorporates a non-linear polynomial data model, site-specific harmonization referenced to a normative site, and variance priors adaptable to small cohorts. It further includes hyperparameter tuning and a goodness-of-fit metric for harmonization assessment. We demonstrate its effectiveness on simulated and real data, showing improved alignment of diffusion metrics and enhanced applicability for normative modeling.
\end{abstract}







\section{Introduction}
\label{sec:introduction}

Neuroimaging studies increasingly use multicenter Diffusion-weighted MRI (DW-MRI) data 
along with normative modeling 
\cite{Di-Biase2023,Marquand2016,Rutherford2023,Verdi2021,Villalon-Reina2022}. Normative modeling is a framework designed to map the general trend and variability of a reference population, facilitating statistical analyses for individual participants with specific diseases. 
The detection of subtle brain variations requires large datasets to ensure robust and sufficient statistical power. However, it is well-established that DW-MRI data is influenced by site-specific acquisition biases, making cross-scanner and cross-protocol comparisons of DW-MRI data unreliable without appropriate data harmonization. Such discrepancies, known as batch effects \cite{Leek2010,Pinto2020}, arise from various technical variability, including software updates, scanner drift, differences in MRI manufacturer, acquisition protocols, field strength, and hardware variability \cite{Cetin-Karayumak2024,Hu2023a,Huynh2019,Moyer2020,Schilling2021}.

The objective of harmonization is to derive a transfer function between two or more sites to eliminate batch effects while preserving covariate expressions, such as age and sex. Although various harmonization methods have been proposed \cite{Chen2022,De-Luca2022,Fortin2017,Horng2022,Hu2023a,Hu2023b,Mirzaalian2016}, many are tailored to specific contexts and may not be suitable for routine clinical practice, which comes with its own set of challenges. For example, data from day-to-day clinical practice often originate from patients with a broad spectrum of diseases and symptoms, unlike clinical studies and academic research, which typically involve carefully selected participants with consistent profiles. Additionally, clinical and academic research protocols are meticulously optimized to achieve the best possible results, with participants spending extended time in scanners to ensure high signal-to-noise ratio (SNR), and scanners across sites usually being from the same vendor. This contrasts with the situation in a network of diverse clinics, where acquisition protocols, clinical practices, and hardware vary significantly.

Currently, the predominant method for DW-MRI harmonization is ComBAT \cite{Fortin2017,Johnson2007}. Unlike other approaches that harmonize the raw diffusion signal \cite{Cetin-Karayumak2019,De-Luca2022,Mirzaalian2016}, ComBAT focuses on harmonizing DW-MRI scalar maps such as Fractional Anisotropy (FA) and Mean Diffusivity (MD) maps \cite{Cetin-Karayumak2020,Fortin2017,Zhu2024}, as well as other features such as cortical thickness and gray matter regions \cite{Fortin2018,Radua2020}. ComBAT employs a Bayesian framework to learn the additive and multiplicative biases of each site, coupled with a linear model that captures the effects of covariates (see Section \ref{sec:method} and the supplementary materials for more mathematical details).

Despite its popularity, ComBAT has limitations. As such, over the years, several derived versions of ComBAT have been proposed to improve the original approach. Most of these improved versions tackle the core assumptions of ComBAT. For instance, ComBAT-GAM \cite{Pomponio2020} counters the linear assumption of ComBAT using spline-based generalized additive models (GAM). M-ComBAT \cite{Da-ano2020} harmonizes data to a specific reference (or target) site rather than a global average. B-ComBAT \cite{Da-ano2020} employs Monte Carlo bootstrapping to robustly estimate ComBAT’s parameters, whereas Longitudinal ComBAT \cite{Beer2020} integrates intra-subject temporal information. CoVBat \cite{Chen2022} additionally corrects for covariance effects besides variance and mean.

While ComBAT is effective in clinical trials and academic research settings, it is not directly applicable to everyday clinical practice as documented by~\cite{Jodoin2025}. In these settings, participants (human or otherwise) undergo specific interventions according to a predetermined research plan \cite{Chew2020}, with data collected at multiple sites following strict protocols. Moreover, they may also encompass retrospective research on public datasets \cite{Aisen2010,Yue2013,Shafto2014,Menze2015,Sudlow2015,Bakas2017,Taylor2017, Jernigan2018,Nugent2024}. ComBAT generally harmonizes all sites simultaneously, an approach that suits a fixed number of sites but falls short in a dynamic clinic network where the number of participants and acquisition sites expand over time. Additionally, ComBAT presupposes a linear distribution of data with respect to its covariates, which is often not the case \cite{Jodoin2025}. Moreover, inter-site variability in regression parameters, such as differences in age-related effects, cannot be captured by the single regression model used in ComBAT \cite{Jodoin2025}. This limitation may cause the ComBAT's model to generalize poorly to unseen data, particularly when trained on a small subject sample or when the subjects cover only a narrow range of covariate values (e.g. subjects with limited age range) \cite{Jodoin2025}.

These limitations necessitate a specialized harmonization approach for clinical settings, which we introduce as Clinical-ComBAT, a mathematical reformulation of the original ComBAT method. 
Clinical-ComBAT builds on years of practical experience harmonizing DW-MRI data across hundreds of heterogeneous sites, where direct application of ComBAT proved impractical. The method addresses these challenges through a simplified data model, polynomial modeling of data, and site-specific harmonization using a normative reference site. Like M-ComBAT~\cite{Da-ano2020} and Pairwise-ComBAT~\cite{Jodoin2025}, it aligns each site with a well-populated reference, avoiding complications associated with increasing site numbers. The large normative population enables integration of new priors during model fitting, accurate estimation of the polynomial order, and the use of goodness-of-fit metrics to assess harmonization quality.

This study compares Clinical-ComBAT with ComBAT \cite{Fortin2017} across four datasets, including three real-world cohorts and one synthetic dataset. Harmonization was applied to the mean diffusivity (MD) and the fractional anisotropy (FA), obtained from Diffusion Tensor Imaging (DTI), and to apparent fiber density \cite{Raffelt2012} derived from the fiber Orientation Distribution Function (fODF) obtained from Constrained Spherical Deconvolution \cite{Descoteaux2007, Tournier2007}. Performance was evaluated across 42 white matter bundles from the IIT Human Brain Atlas v.5.0 and the white matter skeleton mask \cite{Qi2021}. 

The main advances of Clinical-ComBAT over ComBAT can be summarized as follows:
\begin{enumerate}
\item introduction of a new data formulation model,
\item integration of a non-linear polynomial basis function at the core of the model,
\item adoption of a revised Bayesian framework for estimating harmonization parameters,
\item use of a quality control harmonization method
\item implementation of an automatic hyperparameter tuning strategy.
\end{enumerate}
All these contributions are summarized into four algorithms (c.f. the Supplementary Materials) and a ready-to-use publicly available code.

\section{Methods}\label{sec:method}

\subsection{ComBAT fundamentals}
ComBAT is a harmonization method initially devised for genomics to counter batch effects \cite{Johnson2007}. Its effectiveness in mitigating MRI acquisition effects in scalar maps from DW-MRI was subsequently demonstrated \cite{Fortin2017}. Despite its widespread use, the theoretical underpinnings and implementation of ComBAT can be perplexing. 

Let $Y=\{Y_1,Y_2,...,Y_I\}$ represent a set of $I$ physical MRI sites, where $Y_i=\{Y_{i1},Y_{i2},...,Y_{iJ_i}\}$ are a set of $J_i$ participants whose brain has been scanned with the  MRI machine $i$. In voxel-wise harmonization, these participant’s MR images are usually non-linearly registered to a common space, such as the MNI space \cite{Fortin2017}. It's important to note that the number of participants $J_i$ is site dependent, as their number varies from one site to another. Each scalar map $Y_{ij}$ can be expressed as $Y_{ij}=[y_{ij1}, y_{ij2}, ..., y_{ijv}]$, where $v$ is the number of voxels (or region) in the common space and $y_{ijv} \in \mathbb{R}$.

To mitigate site-specific biases, ComBAT employs a linear model for the data formation of each voxel (or regions) $v$ as follows:
\begin{equation}
\label{eq:combat}
    y_{ijv} = \alpha_v + \vec x_{ij}^T \vec \beta_v+ \gamma_{iv} + \delta_{iv}\epsilon_{ijv},
\end{equation}
where $\alpha_v$ is the model intercept of the overall population across all sites, $x_{ij}$ is a vector of covariates, $\beta_v$ is the regression coefficient vector, $\gamma_{iv}$ and $\delta_{iv}$ are the additive and multiplicative effects of site $i$, and $\epsilon_{ijv}$ is a random independent Gaussian noise with $\epsilon_{ijv} \sim \mathcal{N}(0,\sigma_v^2 )$. 
It is crucial to recognize that ComBAT assumes the regression vector $\beta_v$ (i.e. the slope of the population) is constant for all sites. 
The primary goal of ComBAT is to eliminate the site-specific additive and multiplicative biases (respectively $\gamma_{iv}$ and $\delta_{iv}$) ensuring that the harmonized population profile conforms to the model 
\begin{equation}
    \label{eq:combat2}
    y^{ComBAT}_{ijv} = \alpha_v + \vec x^T_{ij}\vec \beta_v + \epsilon_{ijv}.
\end{equation}

Since the parameters of the model $\alpha_v$, $\beta_v$, $\gamma_{iv}$ and $\delta_{iv}$ are a priori unknown, they must be empirically estimated from the data $Y$.  Whenever these parameters can be accurately estimated (namely $\hat{\alpha}_v$, $\hat{\beta_v}$, $\hat{\gamma}_{iv}$, and $\hat{\delta}_{iv}$), the harmonized values $y_{ijv}^{harm}$ can be computed as follows:
\begin{equation}
    y^{harm}_{ijv} = \frac{y_{ijv}-\hat{\alpha}_v - \vec x^T_{ij} \hat{\beta}_v - \hat{\gamma}_{iv}}{\hat{\delta}_{iv}} + \hat{\alpha}_v + \vec x^T_{ij}  \hat{\beta}_v.
\end{equation}

A straightforward approach to estimating these five parameters is the Location and Scale (L/S) method \cite{Johnson2007} that we detailed in section S.1 of the Supplementary Material. 
While effective, this method has notable limitations in presence of small populations. These are typically mitigated by the Bayesian formulation of ComBAT whose mathematical details can be found in section S.2 of the Supplementary Material as well as in previous publications ~\cite{Jodoin2025,Johnson2007,Fortin2017}.

\begin{figure}[!tbh]
\centering
\includegraphics[width=0.95\textwidth]{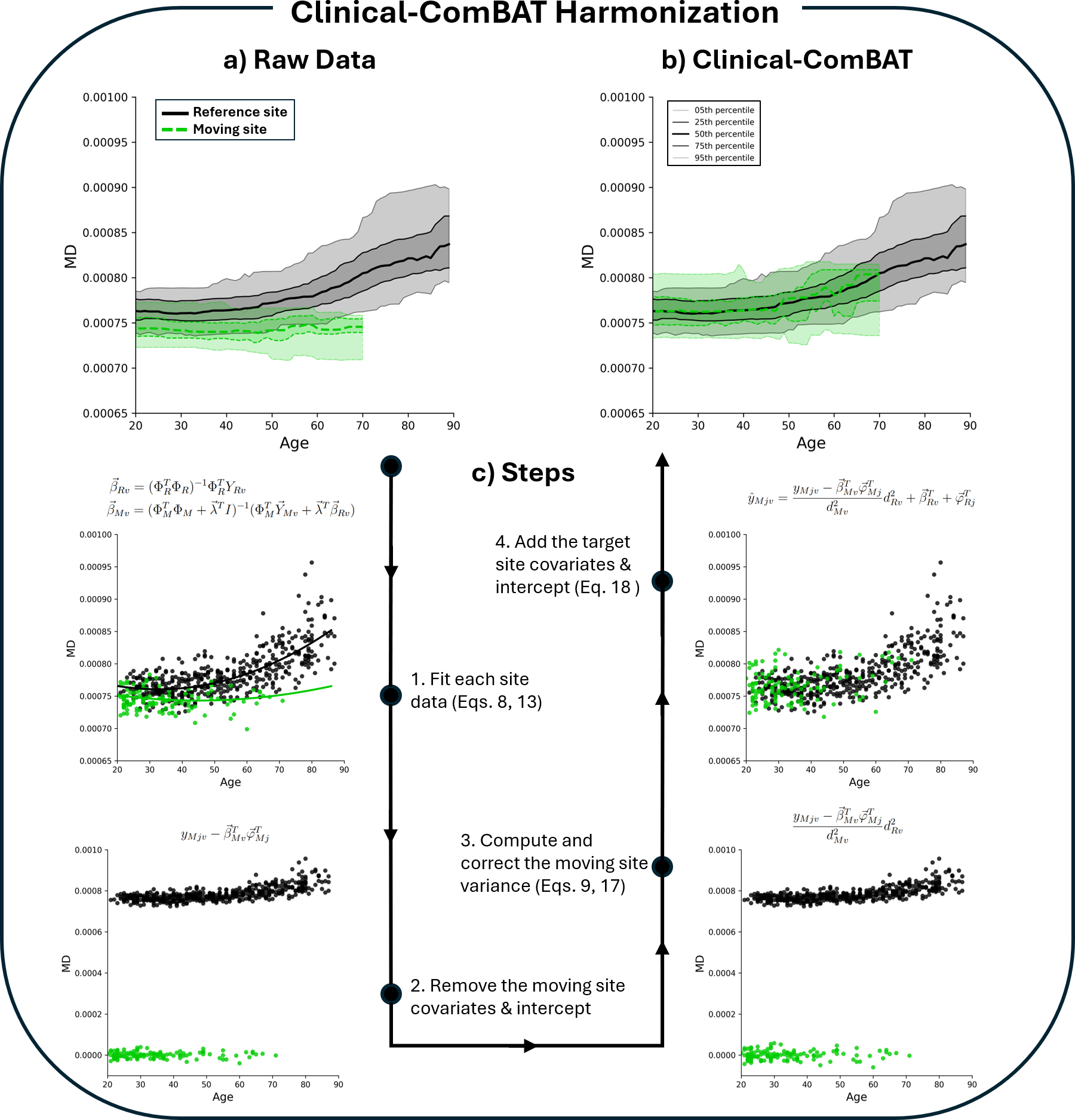}
\caption{Clinical-ComBAT harmonization process for aligning a moving site to a reference site. a) Raw data distributions from the reference (black) and moving (green) sites. b) data after Clinical-ComBAT harmonization. 
c) Step-by-step illustration of the Clinical-ComBAT procedure. Scatter plots show the reference (black) and moving (green) data. Step 1 consists of fitting a regression model to both the reference and moving site data following Eq.(\ref{eq:betaRv}) and (\ref{eq:betaMv}). In Step 2, the site-specific covariate effects and intercept are removed from the moving site data. Step 3 applies a variance correction to align the data dispersion with that of the reference site. In Step 4, the adjusted data are transformed to match the reference site distribution.  Steps 2,3, and 4 derive from Eq.(\ref{eq:Clinical-ComBAT_data}).}\label{fig:Clinical-ComBAT-step}
\end{figure}

\subsection{Clinical-ComBAT Data Formation Model}
Clinical-ComBAT introduces several key modifications to the original ComBAT framework, specifically designed to better address the requirements of clinical practice. The reader shall refer to Figure~\ref{fig:Clinical-ComBAT-step} for a step-by-step illustration of Clinical-ComBAT.  

Its primary objective is to enable reliable comparison of patient data against a normative model, a cornerstone of clinical applications. A distinctive feature of Clinical-ComBAT lies in its strategy of harmonizing each site with a designated reference site. Similar sitewise harmonization strategies have been proposed in methods such as M-ComBAT \cite{Da-ano2020} and Pairwise-ComBAT \cite{Jodoin2025}. In this context, harmonization reduces to aligning the data from a moving site, $D_{Mv} = \{(y_{M1v}, \vec{x}_{M1}),  \dots,$ $(y_{MNv}, \vec{x}_{MN}) \}$, with that of a reference site, $D_{Rv} = \{(y_{R1v}, \vec{x}_{R1}), \dots,$ $(y_{RNv}, \vec{x}_{RN}) \}$. Clinical-ComBAT assumes that the reference site contains a sufficiently large and diverse healthy control (HC) population to establish a robust normative model. This assumption relies on the diversity of the reference cohort, which should span a wide age range, include both handedness profiles, and ensure a balanced representation of genders.

The sitewise harmonization approach in Clinical-ComBAT addresses two challenges. First, it avoids the complexity of needing to harmonize an ever-growing number of sites simultaneously, as each site is harmonized independently. Second, it resolves the problem with ComBAT, which could inadvertently modify the harmonization parameters for already calibrated sites when applied indiscriminately to a new site.

Additionally, sitewise harmonization simplifies the data formation equation used in ComBAT (c.f. Eq.~\ref{eq:combat}). At the core of most ComBAT methods is the assumption that each dataset includes two additive biases (a population bias $\alpha_v$ and a site bias $\gamma_{iv}$) and two multiplicative biases (a population bias $\sigma_v$ and a site bias $\delta_{iv}$). These four parameters are intended to represent the variations of each site from an overall unbiased reference (yet normative) population. 
Since Clinical-ComBAT independently harmonizes each moving site to a reference site, the data formation model can be streamlined to two biases. To do so, Clinical-ComBAT combines the $\alpha_v+\gamma_{iv}$ additive biases of Eq.~(\ref{eq:combat}) into a single sitewise additive bias, denoted as $b_{iv}$, where $i\in \{M,R\}$ is the index for the moving (M) and the reference (R) site. Additionally, because the multiplication of a normally distributed variable $\epsilon_{ijv} \sim \mathcal{N}(0,\sigma_v^2)$ by $\delta_{iv}$ still results in a normal distribution with variance $(\sigma_v\delta_{iv})^2$, our model adopts a single multiplicative bias, denoted as $d_{iv}$. Compared to  Eq.~(\ref{eq:combat}), these modifications lead to a simplified sitewise data formation model:
\begin{equation}
    \label{eq:Clinical-ComBAT}
    y_{ijv} = b_{iv} + \vec x_{ij}\vec \beta_v + d_{iv}\epsilon,
\end{equation}
where $i \in \{M,R\}$ and $\epsilon_{ijv} \sim \mathcal{N}(0,1)$. 

Moreover, Clinical-ComBAT introduced nonlinearity in the model. This nonlinearity is achieved using a polynomial basis function of positive degree $P$, expressed as
\begin{equation}
    \label{eq:phi}
    \varphi(\Vec{x_{ij}}) = (\vec{x}^T_{ij} \cdot \vec{x}_{ij} + 1)^P,
\end{equation}
denoted as $\vec{\varphi}_{ij}$ for notation simplicity. Notably, regardless of the degree $P$, the expansion of this polynomial always includes a constant term, which allows the additive bias $b_{iv}$ to be effectively integrated into it. Consequently, the data formation equation is revised to:
\begin{equation}
    \label{eq:data_formation}
    y_{ijv} = \vec{\varphi}^T_{ij}\vec{\beta}_{iv} + d_{iv}\epsilon_{ijv}.
\end{equation}
Here, each site is characterized by a specific set of weights $\vec{\beta}_{iv}$, along with a single multiplicative bias $d_{iv}$ and an additive bias that is incorporated into the $\vec{\beta}_{iv}$ vector. 

\subsection{Harmonization Fit}
With the sitewise harmonization framework of Clinical-ComBAT and according to Eq.~(\ref{eq:data_formation}), the harmonization of a moving site $M$ onto a target normative reference $R$ requires the estimation of four terms: two weight vectors $\vec{\beta}_{Mv}$ and $\vec{\beta}_{Rv}$ and two multiplicative biases $d_{Mv}$ and $d_{Rv}$. 

\paragraph{Reference site parameters $\vec{\beta}_{Rv}, d^2_{Rv}$}
$\vec{\beta}_{Rv}$ is computed like ComBAT, by maximizing a likelihood Gaussian distribution of the reference data. Doing so boils down to minimizing the following L2 loss function \cite{Bishop2006,Duda2001}
\begin{equation}
    \vec{\beta}_{Rv} = \arg\max_{\vec{\beta}} \frac{1}{J_R} \sum_{j=1}^{J_R} (y_{Rjv} - \vec\varphi_{Rj}^T\vec\beta_{Rv})^2,
\end{equation}
that one can solve with a close form solution:
\begin{equation}
    \label{eq:betaRv}
    \vec{\beta}_{Rv} = (\Phi_R^T\Phi_R)^{-1} \Phi_R^T Y_{Rv},
\end{equation}
where $\Phi_R$ is a matrix containing the concatenated covariate vectors $\vec\varphi_{Rj}$ of all reference participants, and $Y_{Rv}$ is a vector containing the values $y_{Rjv}$ of the reference site. While this formulation resembles the L/S ComBAT method \cite{Johnson2007}, Clinical-ComBAT relies solely on the reference site data, whereas  L/S ComBAT incorporates data from all sites.

Similarly, with the hypothesis that the reference site is well populated, the estimation of $d^2_{Rv}$ does not require a Bayesian prior as in ComBAT \cite{Fortin2017}. Instead, $d^2_{Rv}$ can be estimated following a maximum likelihood, which is a simpler alternative to ComBAT’s iterative solution required by the use of an inverse gamma prior (c.f. Supplementary Material S.2). Instead, $d^2_{Rv}$ is obtained following a maximum likelihood of a Gaussian distribution which leads to the calculation of the simple variance of the rectified data \cite{Duda2001}, i.e.
\begin{equation}
\label{eq:varrefsite}
    d^2_{Rv} = \frac{1}{J_R} \sum_{j=1}^{J_R}(y^T_{jv} - \vec\varphi^T_{Rv}\vec\beta_{Rv})^2.
\end{equation}

\paragraph{Moving site parameters $\vec{\beta}_{Mv},  d^2_{Mv}$} Since the moving site may contain a limited number of data, we proceed with the estimation of $\vec\beta_{v}$ and $d_{Mv}$ with a maximum a posteriori (MAP) formulation. Unlike ComBAT which models $\vec\beta_{Mv}$ with a likelihood distribution, Clinical-ComBAT models it with a posterior distribution involving the product of a Gaussian likelihood and prior:
\begin{equation}
    P(\vec\beta_{Mv} | D_{Mv}) \propto P(D_{Mv} | \vec\beta_{Mv})P(\vec\beta_{Mv})
\end{equation}
where $D_{Mv} = \{Y_{Mv}, X_M \}$ contains the list of all values and covariates of the moving site at location $v$  (e.g. voxel or region) and where
\begin{equation}
    P(D_{Mv}|\vec\beta_{Mv}) = \mathcal{N}(\Phi^T_M\vec\beta_{iv}, \Sigma_l),
\end{equation}
\begin{equation}
    P(\vec\beta_{Mv}) = \mathcal{N}(\vec\beta_{iv}, \Sigma_0).
\end{equation}
The goal of the prior is to make sure $\vec\beta_{Mv}$ does not differ too much from $\vec\beta_{Rv}$, the regression weights of the reference site. This is critically important when a moving site accounts for a low number of participants. Maximizing the posterior $P(\vec\beta_{Mv} | D_{Mv})$ leads to the following close-form solution:
\begin{equation}
    \label{eq:betaMv} 
    \vec\beta_{Mv} = (\Phi^T_M\Phi_M + \vec\lambda^TI)^{-1}(\Phi^T_M\vec{Y}_{Mv} + \vec\lambda^T\vec\beta_{Rv}),
\end{equation}
where $I$ is the identity matrix and $\vec\lambda \in \mathbf{R}^{p+}$ is an hyperparameter regularization vector -please refer to section S3 of the Supp mat for the proof.  

Eq(\ref{eq:betaMv}) comes with an intuitive flavor. For $\vec\lambda = \vec0$, the regression of $\vec\beta_{Mv}$ is a maximum likelihood estimation as in Eq.~(\ref{eq:betaRv}). As the regularization increases, more weight is put on $\vec\beta_{Rv}$, and, when $\vec\lambda$ is large enough, $\vec\beta_{Mv} \approx \vec\beta_{Rv}$.

Clinical-ComBAT estimated $d^2_{Mv}$ with a posterior given by
\begin{equation}
    P(d^2_{Mv} | z_{Mv}) \approx P(z_{Mv}|d^2_{Mv})P(d^2_{Mv}),
\end{equation}
where $z_{Mv}$ is the set of all rectified values of the moving site, namely $z_{Mv} = y_{Mjv} - \vec\varphi_{Mj} \vec\beta_{Mv}\; \forall j$. Like ComBAT, the likelihood distribution follows a Gaussian distribution. However, Clinical-ComBAT uses a prior Gamma distribution (instead of an inverse Gamma for ComBAT) :
\begin{eqnarray}
    P(z_{Mv} | d^2_{Mv}) = \mathcal{N}(0,d^2_{Mv}) \\
    P(\lambda_{Mv}) = \mathcal{G}(\lambda_{Mv},a_0,b_0),
\end{eqnarray}
where $\lambda_{Mv} = \frac{1}{d^2_{Mv}}$ is the inverse of the variance \cite{Gelman2013}. By incorporating the variance of the reference site $d^2_{Rv}$ in the Gamma prior, maximizing the posterior leads to the following solution:
\begin{equation}
    \label{eq:nu}
    d^2_{Mv} = \frac{J_M \hat d^2_{Mv}}{J_M + \nu} + \frac{\nu d^2_{Rv}}{J_M + \nu},
\end{equation}
where $J_M$ is the number of data in the moving site, $\hat d^2_{Mv}$ is the empirical variance of the rectified moving site: $\frac{1}{J_M} \sum^{J_M}_{j=1}(y_{Mjv}-\vec\varphi^T_{Mj} \vec\beta_{Mv})^2$ and $\nu \in \mathbf{R}^+$ is a hyperparameter -c.f. Section S.3 of the Supplementary Material for the proof.

This equation comes with a strikingly intuitive flavor. When the moving site lacks data, indicated by $J_M=0$, the variance of the moving site $d_{Mv}^2$ defaults to match the previously-estimated variance of the reference site $d_{Rv}^2$. As the quantity of data of the moving site increases, $d_{Mv}^2$ transitions to a weighted average between $\hat d_{Mv}^2$ and $d_{Rv}^2$. At some point, if $J_M = \nu$, $d_{Mv}^2$ becomes the exact average between $\hat d_{Mv}^2$ and $d_{Rv}^2$. Beyond this point, when $J_M \gg \nu$, $d_{Mv}^2$ closely approximates to $\hat d_{Mv}^2$. In addition to its simplicity, Eq.~(\ref{eq:nu}) does not require an iterative process as in ComBAT and has the reference site variance as a prior instead of data taken across other regions of the brain. 

Please note that the procedure for estimating the four(4) harmonization parameters $\{d_{Mv}^2, \vec\beta_{Mv},d_{Rv}^2, \vec\beta_{Rv} \}$ is summarized in the \texttt{Clinical-ComBAT-Fit} algorithm presented in section S.4 of the Supplementary Material.

\subsection{Harmonization Process}
Once the four parameters $\{d_{Mv}^2, \vec\beta_{Mv},d_{Rv}^2, \vec\beta_{Rv} \}$ have been estimated, the data from the moving site can be harmonized with the reference site as follows:
\begin{equation}
    \label{eq:Clinical-ComBAT_data}
    \hat{y}_{Mjv}^{C-ComBAT} = \frac{y_{Mjv} - \vec\beta^T_{Mv}\vec{\varphi}_{Mj}}{d^2_{Mv}}d^2_{Rv} +  \vec\beta^T_{Rv} + \vec{\varphi}^T_{Rj}.
\end{equation}

Figure \ref{fig:Clinical-ComBAT-step} depicts step-by-step the harmonization procedure of Clinical-ComBAT. Together, these steps produce a harmonized dataset in which the moving site is aligned with the reference site.
The harmonization procedure is summarized in the \texttt{Clinical-ComBAT-Apply} algorithm presented in section S.4 of the Supplementary Material.

\subsection{Goodness-of-Fit Quality Control}

To assess harmonization quality, we compute a population overlap score, measuring the degree of overlap between the data of the reference site $D_{Rv}$ and that of the harmonized moving dataset $\hat D_{Mv}$.  This is achieved by rectifying both populations using the reference parameter vector $\vec\beta_{Rv}$ (c.f. Fig.~\ref{fig:Clinical-ComBAT-step} Step 2 for an illustration of the rectification of the moving site):
\begin{eqnarray}
    z_{Rjv} &=& y_{Rjv} - \vec\varphi^T_{Rj} \vec\beta_{Rv} \;\;\;\;\;\;\;\forall (y_{Rjv}, \vec\varphi^T_{Rj}) \in D_{Rv}, \\
    \hat z_{Mjv} &=& \hat y_{Mjv} - \vec\varphi^T_{Mj} \vec\beta_{Rv} \;\;\;\;\;\forall (\hat y_{Mjv}, \vec\varphi^T_{Mj}) \in D_{Mv}.
\end{eqnarray}

This rectification process, by removing the effect of the covariates, allows assuming that $z_{ijv}$ is independent of the data vector $\vec \varphi_{ij}$ and that $P(z_{Rjv} | \vec \varphi_{Rj}) = P(z_{Rjv})$ and $P(\hat z_{Mjv} | \vec \phi_{Mj}) = P(\hat z_{Mjv})$ two Gaussian distributions following the ComBAT assumption.  

The distance between univariate Gaussian distribution functions is computed using the Bhattacharyya distance $d_B$ \cite{Jodoin2025}:

\begin{equation}
d_B = \frac{1}{4} \frac{(\mu_R-\mu_M)^2}{\sigma^2_R + \sigma^2_M} + \frac{1}{2} \ln (\frac{\sigma^2_R + \sigma^2_M}{2\sigma_R\sigma_M}),
\end{equation}
where $\mu_R, \mu_M$ are the mean and $\sigma_R, \sigma_M$ the standard deviation of the rectified reference and moving sites. The goodness-of-fit QC procedure is summarized in the \texttt{Harmonization-QC} algorithm in section S.4 of the Supplementary Material.

\subsection{Hyperparameter Auto-tuning}
Clinical-ComBAT introduces a hyperparameter vector $\vec\lambda$ in Eq.(\ref{eq:betaMv}), which controls the degree to which the moving-site weights $\vec\beta_{Mv}$ deviate from the reference-site weights $\vec\beta_{Rv}$. At one extreme, when $\vec\lambda \rightarrow \vec\infty$, the moving-site weights converge to those of the reference site ($\vec\beta_{Mv} \approx \vec\beta_{Rv}$). At the other extreme, when $\vec\lambda \rightarrow \vec 0$, $\vec\beta_{Mv}$ aligns closely with the moving-site data. An excessively large $\vec\lambda$ may result in underfitting, whereas an overly small $\vec\lambda$ can lead to overfitting, particularly when the polynomial degree $P$ in Eq.(\ref{eq:phi}) is high, or when the moving dataset is small, includes outliers, or spans a narrow age range \cite{Jodoin2025}.   

\begin{figure}[t]
\centering
\includegraphics[width=0.9\textwidth]{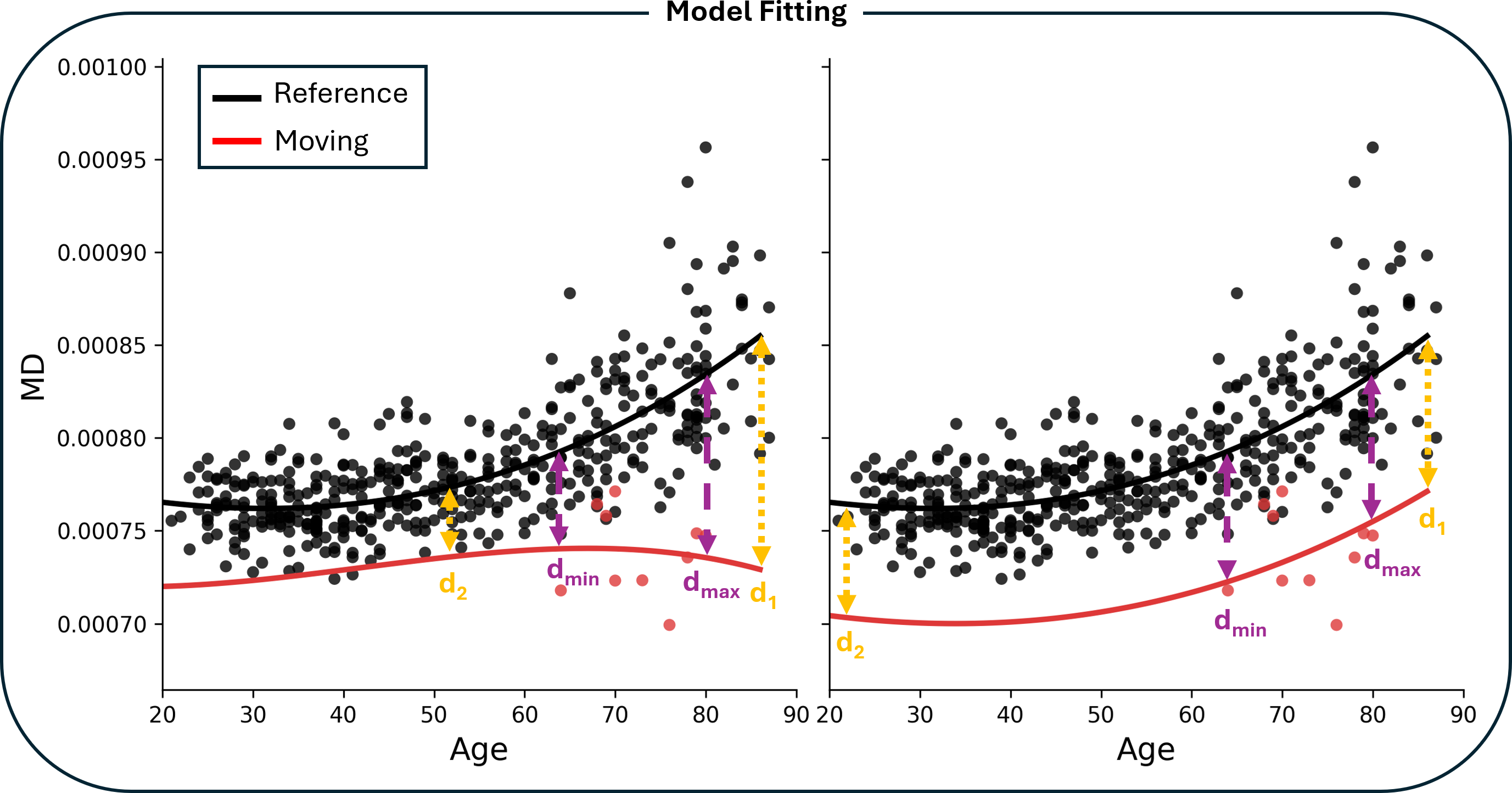}
\caption{Example of two model fits using a third-degree polynomial ($P=3$). Black dots and their fitted curve correspond to the reference Cam-CAN dataset, while red dots represent 10 synthetic moving data points. Left: the moving model with $\vec{\lambda}=0$ overfits the data and diverges outside the 60–80 age range. Right: the auto-tuned model ($\vec{\lambda}=32762$) does not deviate too much from the reference model.}\label{fig:hyperparameter}

\end{figure}

The goal of the auto-tuning is to identify the appropriate $\vec\lambda$ vector that ensures the moving curve aligns closely with the reference curve while accurately representing the moving data. As illustrated in Figure~\ref{fig:hyperparameter}, it uses four distance measures between the two curves. The first two measures, $d_{min}$ and $d_{max}$ represent the minimum and maximum distances between the curves across the age values present in the moving data:
\begin{eqnarray}
    d_{min} &=& |\min \vec\beta^T_{Rv} \vec\varphi_j - \vec\beta^T_{Mv} \vec\varphi_j| \;\;\;\forall j x_M, \\
    d_{max} &=& |\max \vec\beta^T_{Rv} \vec\varphi_j - \vec\beta^T_{Mv} \vec\varphi_j|\;\;\; \forall j x_M.
\end{eqnarray}
Here, $x_M$ denotes the set of covariates at the moving site. The other two distances, $d_1$ and $d_2$, are the minimum and maximum inter-curve distances across the full age range.  

The optimization of $\vec\lambda$ can be expressed as follows:
\begin{equation}
    \label{eq:lambda}
    \vec\lambda = \arg\min_{\vec\lambda} sign(d_{\min} \frac{1}{\tau} - d_1) + sign(d_2 - d_{\max} \tau) + 2,
\end{equation}
where $\tau \geqq 1$ is a predefined value, and the $sign(x)$ function is defined as 1 if $x \geq 0$, -1 otherwise.
Note that when $\tau=1$, the optimized $\vec\lambda$ vector results in a moving curve perfectly parallel to the reference curve. 
A suitable $\vec\lambda$ vector can be found by incrementally increasing its value until the conditions of Eq.~(\ref{eq:lambda}) are satisfied. 

The auto-tuning procedure is summarized in the \texttt{Hyperparameter auto-tuning} is section S.4 of the Supplementary Material. 

\subsection{Data and Processing}
\label{sec:dataset}
Experiments comparing Clinical-ComBAT to ComBAT for DW-MRI-derived metric harmonization were conducted on four datasets.  The first one is the Cambridge Centre for Ageing Neuroscience (CamCAN) dataset \cite{Shafto2014,Taylor2017}, which serves as the reference site. All reference subjects were acquired at the same MRI machine (Siemens 3T TIM Trio MRI machine). The 441 HC, aged 18–87 years old, correspond to the subset of subjects acquired before the scanner upgrade. The DW-MRI protocol includes 60 gradient directions evenly distributed across two b-values (1000 and 2000 $s/mm^2$ and three non-diffusion-weighted measurements, with a $2\times 2\times 2$ $mm^3$ image resolution. The second datasets are National Institute of Mental Health (NIMH) Intramural Healthy Volunteer Dataset~\cite{Nugent2022}. The NIMH dataset is composed of 119 HC, aged 18–71 years old. The DW-MRI data were acquired at $b=1,000 s/mm^2$ for 48 uniformly distributed gradient directions ($2\times 2\times 2~mm^3$, General Electric 3T MRI machine). The third and forth dataset are from the Track-TBI Network \cite{McMahon2014} (site A and site B) with 55 HC and 104 Traumatic brain injury (TBI) patients, aged 19-68. The DW-MRI protocol included 64 $b=1,300s/mm^2$, with the image resolution set to $2.7\times 2.7\times 2.7~mm^3$. Site A (University of Washington) was acquired on the Philips 3T MRI machine, and Site B (University of Texas at Austin) was acquired on a Siemens 3T MRI machine. The NIMH and Track-TBI dataset were used as moving sites.

The image processing was performed using the Tractoflow pipeline \cite{Theaud2020,Jenkinson2012,Garyfallidis2014} for all subjects. T1-weighted images were registered to both the MNI template \cite{Fonov2009} and to DW-MRI data using ANTs \cite{Avants2009}. Diffusion Tensor Imaging (DTI) maps (FA, MD) and the apparent fiber density map \cite{Raffelt2012}, derived from the fiber Orientation Distribution Function (fODF) obtained from Constrained Spherical Deconvolution \cite{Descoteaux2007, Tournier2007}, were registered to the MNI template using the computed transformations. DW-MRI images with b-values below $b=1200 s/mm^2$ were used to compute the Diffusion Tensors, and b-values above $b=700 s/mm^2$ were used to compute the fODFs. The fODFs were generated using a spherical harmonics order of $8$ and a standardized response function~\cite{Descoteaux2007} for all subjects ($15, 4, 4$) x $10 ^4\mathrm{ms}/\mathrm{\mu m}^2$. The IIT Atlas (v.5.0) \cite{Qi2021} was aligned to the MNI template \cite{Avants2009}, and mean MD, FA, and AFD values for each of the 42 bundles and the white matter skeleton mask were computed using the streamline density map.

The last dataset is called {\em Modified-CamCAN}~\cite{Jodoin2025}. It is a synthetic dataset based on the 441 healthy controls from the original CamCAN cohort, with altered mean diffusivity (MD) values. Controlled additive and multiplicative biases were introduced according to:
\begin{equation}
\label{eq:combat_M1_M2_A_Bias}
y_{ijv} = A\cdot b_{iv} + S\cdot \vec x_{ij}\vec \beta_v + M\cdot d_{iv}\epsilon
\end{equation}
where $A\in R$ is a shift in the mean, $S\in R$ modulates covariate effects (slope), and $M\in R$ scales the residual variance \cite{Jodoin2025}.

\section{Results}

This section presents a series of four experiments.
Unless otherwise mentioned, through these experiments we used a prior $\nu$ of 5 for the estimation of the moving site variance (c.f. Eq.\ref{eq:nu}) a polynomial degree of $P=2$ (c.f. Eq.(\ref{eq:phi})) and a value of $\tau = 2$ for the auto-tuning (c.f. Eq.\ref{eq:lambda}). All white matter bundles and metrics were independently harmonized.

\subsection{Harmonization Performances} 
\begin{figure}[!tbh]
\centering
\includegraphics[width=0.87\textwidth]{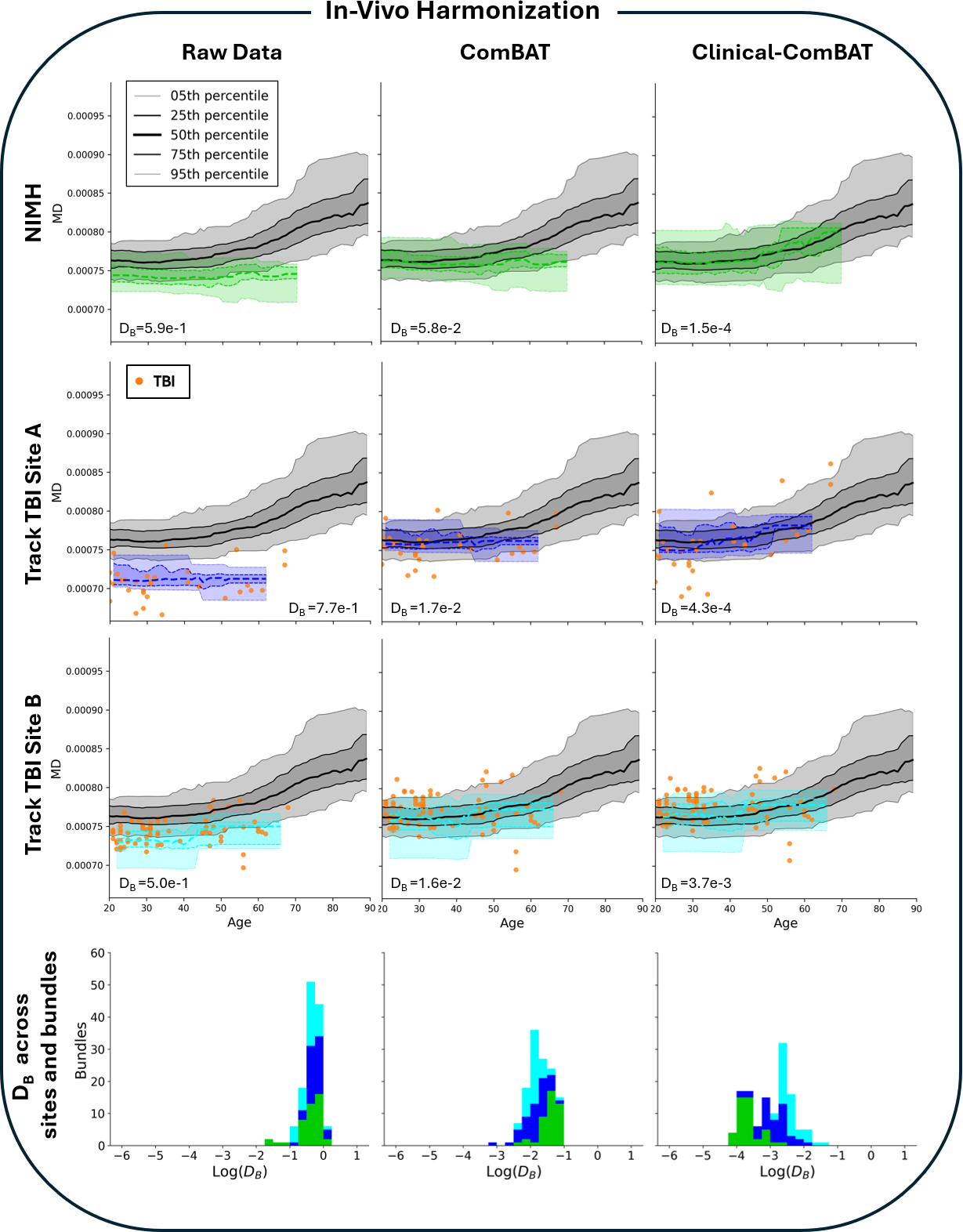}
\caption{Harmonization of mean diffusivity (MD) using Clinical-ComBAT and ComBAT for the NIMH site (green), and Track-TBI site A (blue) and site B (cyan). The top three rows show harmonized white matter skeleton masks with corresponding $D_B$ values, where orange dots indicate TBI subjects. The bottom row presents stacked histograms of $D_B$ across all white matter bundles and sites. Lower $D_B$ values indicate closer alignment with the reference site, underscoring both the necessity of harmonization and the superior performance of Clinical-ComBAT over ComBAT.}\label{fig:results-md}
\end{figure}

Figure \ref{fig:results-md} compares the mean diffusivity (MD) harmonization using Clinical-ComBAT (right), ComBAT (center), and non-harmonized data (left) with a Bhattacharyya distance ($D_B$) value at the bottom left of each plot. The plots of the top three rows illustrate data from moving sites (colored) overlaid on the CamCAN reference site (black) distribution within the white matter skeleton mask. The misalignment of non-harmonized distributions with the reference highlights the necessity of harmonization. ComBAT improves alignment but bias remains. The slope is not properly corrected for the NIMH site for the age range 40 to 60 years old. The multiplicative bias is overcorrected for the Track TBI site A, and undercorrected for Track TBI site B. In contrast, Clinical-ComBAT provides a more accurate bias correction, yielding a distribution closer to the reference site. The bottom row of Figure \ref{fig:results-md} quantifies this improvement, showing $D_B$ histograms for 42 bundles before and after harmonization. While both methods reduce $D_B$, Clinical-ComBAT achieves lower values, enhancing cross-site alignment in healthy controls. Similar trends are observed for Fractional Anisotropy (FA) and Apparent Fiber Density (AFD), shown in Supplementary Figures.

\subsection{Site-Specific Regression Parameters}
\begin{figure}[!tbh]
\centering
\captionsetup[subfigure]{labelformat=empty}%
\subcaptionbox{\label{fig:results-synthetic-md-a}}{}%
\subcaptionbox{\label{fig:results-synthetic-md-b}}{}%
\subcaptionbox{\label{fig:results-synthetic-md-c}}{}%
\subcaptionbox{\label{fig:results-synthetic-md-d}}{}%
\includegraphics[width=1\textwidth]{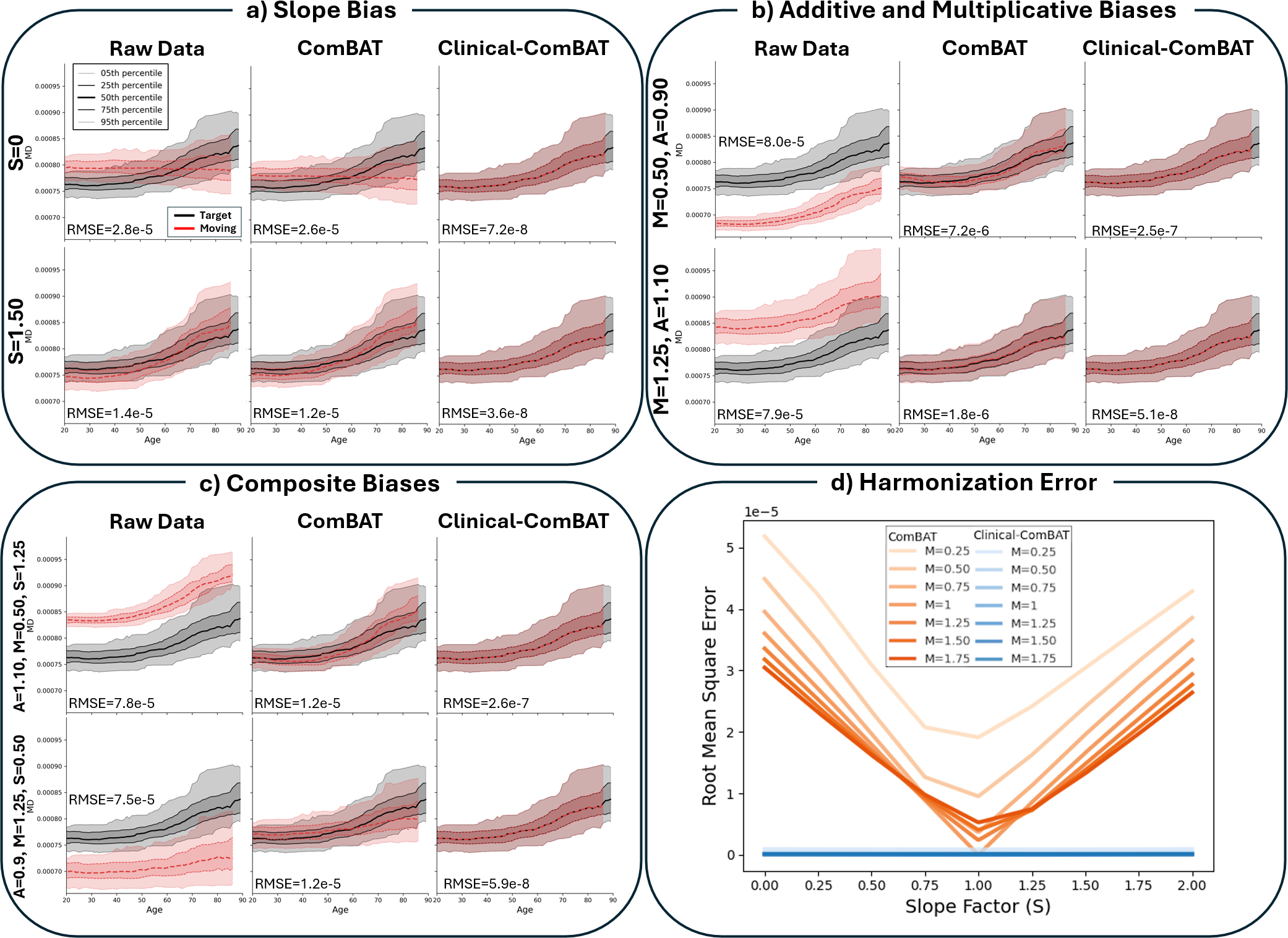}
\caption{Synthetic mean diffusivity (MD) data harmonization performance of ComBAT and Clinical-ComBAT.  
a–c) Distributions of target and moving site data before and after harmonization under different simulated biases: a) slope bias; b) combined multiplicative and additive biases; c) combined slope, multiplicative, and additive biases. d) Root Mean Squared Error (RMSE) of ComBAT and Clinical-ComBAT for increasing slope and multiplicative bias.}\label{fig:results-synthetic-md}
\end{figure}

A key advantage of Clinical-ComBAT is its use of site-specific regression parameters, enabling harmonization in scenarios where the single shared model across site of ComBAT fails. This is demonstrated in Figure \ref{fig:results-synthetic-md}, where the slope of the moving data was artificially varied from 0 (no age effect) to 2 (amplified age effect). It presents the harmonization performance of ComBAT and Clinical-ComBAT on synthetic mean diffusivity (MD) data. Figure~\ref{fig:results-synthetic-md-a}) shows the distributions of target and moving site data before and after harmonization for a modified slope in the moving site. ComBAT fails to adequately correct for slope-induced bias, as shown by high Root Mean Squared Error (RMSE), particularly when the slope diverges from the reference ($S=0$). In contrast, Clinical-ComBAT successfully aligns the moving site data to the reference across all slope conditions. Figures \ref{fig:results-synthetic-md-b}–\ref{fig:results-synthetic-md-c} illustrate harmonization under combined simulated biases. While ComBAT improves cross-site alignment, Clinical-ComBAT yields superior correction, particularly in Figure \ref{fig:results-synthetic-md-c}, where the simulated data include a slope bias. Figure~\ref{fig:results-synthetic-md-d}) summarizes RMSE across a range of slope ($S = [0, 2]$) and multiplicative ($M = [0.25, 1.75]$) biases. Clinical-ComBAT maintains RMSE below $9.4 \times 10^{-7}$ across all conditions, while ComBAT provides its best performances only within $S = [0.75, 1.25]$ and $M = [0.75, 1.50]$, with performance degrading under more pronounced biases.

\subsection{Harmonization of Unseen Data}

Clinical-ComBAT estimates harmonization parameters from the available data (training) of a moving site and applies them to new, unseen subjects (test). This situation is at the core of a day-to-day clinical practice, as new subjects are scanned every day. To address the challenges of limited healthy control data and potential extrapolation beyond the observed covariate range (e.g., age), Clinical-ComBAT incorporates priors from the reference site to regularize both regression parameters and variance estimation.

\subsubsection{Limited Number of Subjects}
\begin{figure}[!tbh]
\centering
\captionsetup[subfigure]{labelformat=empty}%
\subcaptionbox{\label{fig:results-synthetic-md_number-a}}{}%
\subcaptionbox{\label{fig:results-synthetic-md_number-b}}{}%
\includegraphics[width=1\textwidth]{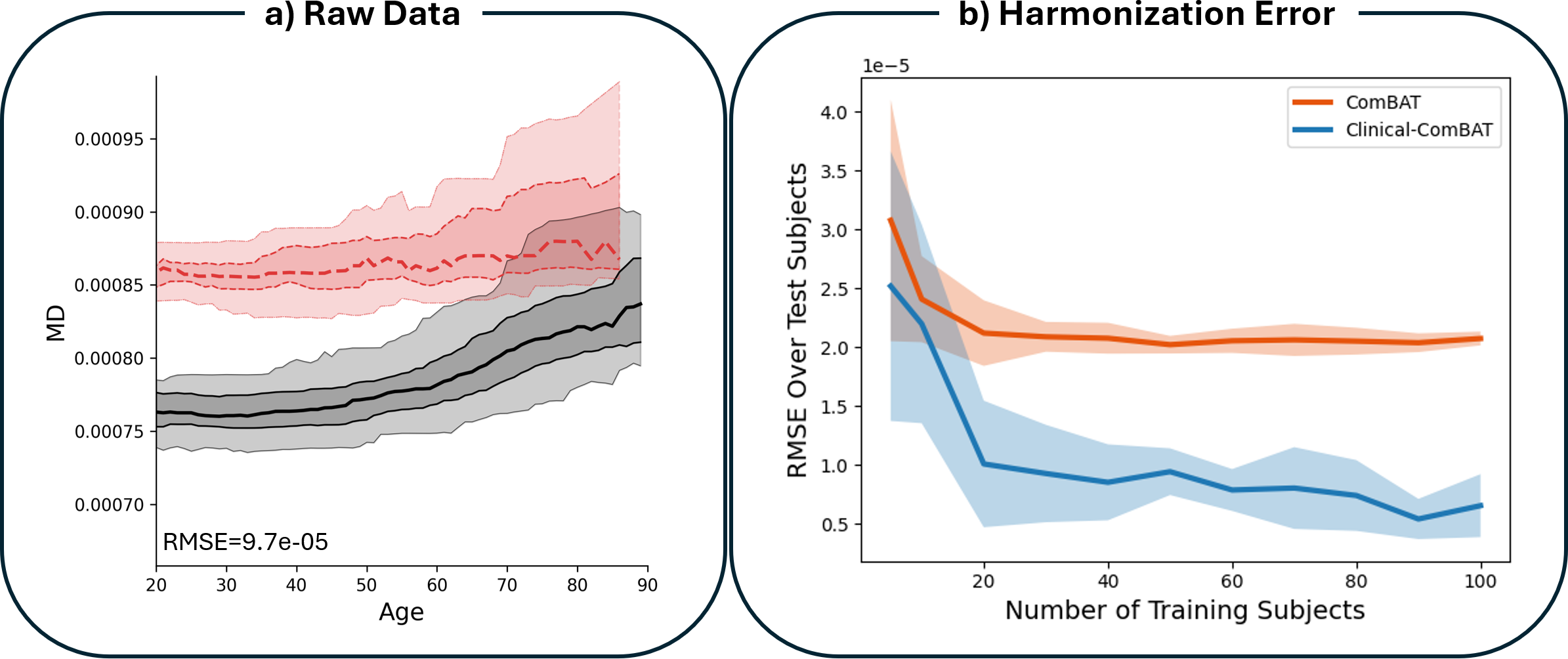}
\caption{Harmonization performance of ComBAT and Clinical-ComBAT on synthetic mean diffusivity (MD) data of the white matter skeleton mask, with limited amount of training data. The test set comprises 100 randomly selected subjects, and the remaining 341 subjects are used to generate training sets. a) Target site data (black) and raw moving site data (red) under simulated biases: slope $S = 0.50$, multiplicative $M = 1.25$, and additive $A = 1.10$. b) Root Mean Squared Error (RMSE) across increasing numbers of randomly sampled training subjects. Each condition was repeated 30 times. The curves show mean RMSE and shaded areas represent one standard deviation.}\label{fig:results-synthetic-md_number}
\end{figure}

Figure \ref{fig:results-synthetic-md_number} illustrates the harmonization performance of ComBAT and Clinical-ComBAT on synthetic mean diffusivity (MD) data under limited number of subjects. Figure \ref{fig:results-synthetic-md_number-a}) shows the target site data (black) and the raw moving site data (red) with simulated slope ($S = 0.75$), multiplicative ($M = 1.50$), and additive ($A = 0.90$) biases. Figure \ref{fig:results-synthetic-md_number-b}) reports RMSE as a function of the number of randomly sampled training subjects selected from a pool of 341 (30 repetitions). The RMSE is computed over 100 unseen test subjects. The curves indicate the mean, and shaded areas represent one standard deviation. By fitting the moving data independently, Clinical-ComBAT manages to outperform ComBAT. This is especially visible when the number of subject is superior to twenty.

\begin{figure}[!tbh]
\centering
\captionsetup[subfigure]{labelformat=empty}%
\subcaptionbox{\label{fig:nbr-sub-a}}{}%
\subcaptionbox{\label{fig:nbr-sub-b}}{}%
\includegraphics[width=1\textwidth]{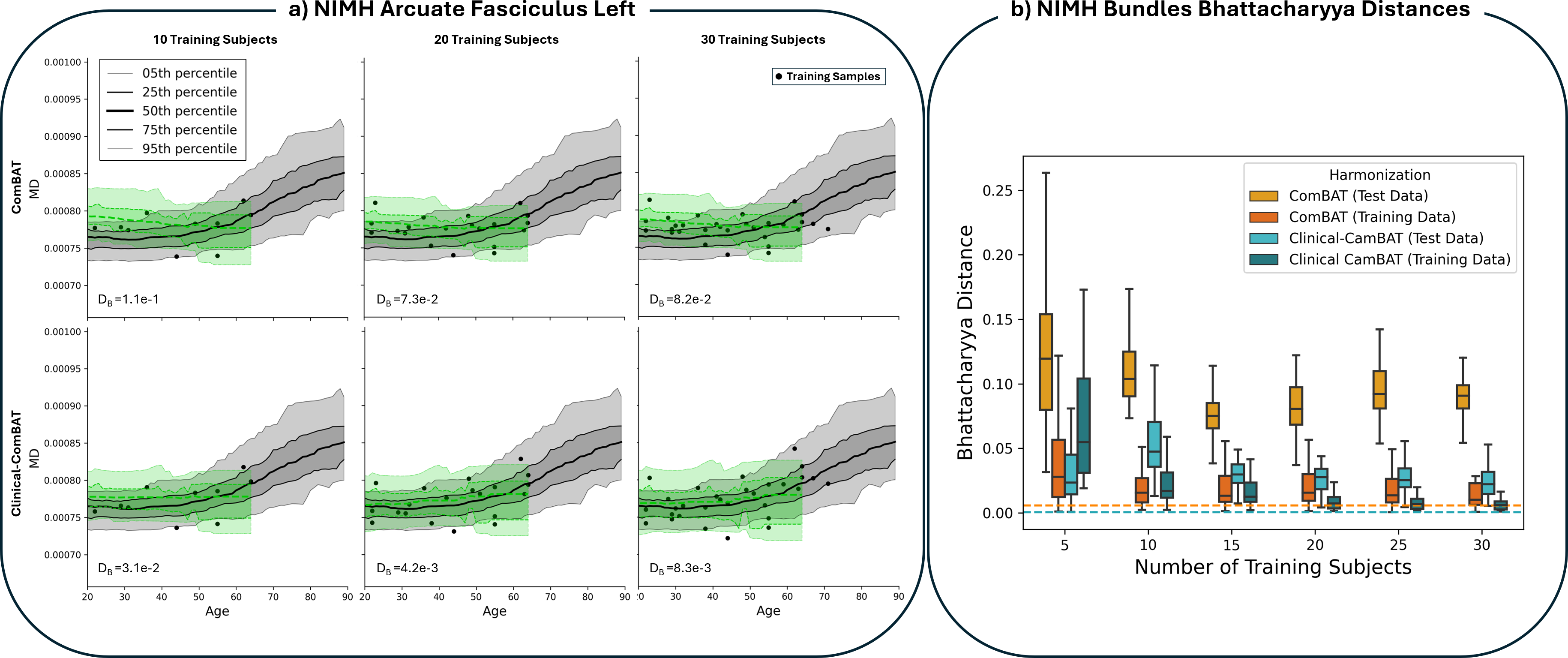}
\caption{Impact of the number of samples on Harmonization (MD). (a) The ComBAT and Clinical-ComBAT harmonized moving site (NIMH, arcuate fasciculus left bundle) data using 10, 20 and 30 samples (black dots). The test data (89 subjects, green percentiles) were harmonized using the model trained on a subset of independent samples. The $D_B$ are reported for the test data. (b) Bhattacharyya distances across all bundles (42) for training samples ranging from 5 to 30. The color dotted lines show the $D_B$ when harmonizing with all data (119 subjects) for ComBAT (orange) and Clinical-ComBAT (cyan).
}\label{fig:nbr-sub}
\end{figure}

Figure \ref{fig:nbr-sub} illustrates the effect of sample size on harmonization performance for ComBAT and Clinical-ComBAT on in-vivo data. In Figure \ref{fig:nbr-sub-a}), the harmonization of mean diffusivity (MD) of the arcuate fasciculus left bundle is shown for 10, 20, and 30 training subjects (black dots) from the NIMH site, and the green percentiles represent the corresponding test set. Clinical-ComBAT yields improved percentile alignment with the reference site, particularly in slope estimation. Figure \ref{fig:nbr-sub-b}) presents the mean and standard deviation of the $D_B$ across 42 bundles for both training and test sets. Clinical-ComBAT consistently achieves lower $D_B$ on test data, indicating more robust harmonization with a limited number of healthy controls.

\subsubsection{Limited Age Range}
\begin{figure}[!tbh]
\centering
\captionsetup[subfigure]{labelformat=empty}%
\subcaptionbox{\label{fig:results-synthetic-md_age-a}}{}%
\subcaptionbox{\label{fig:results-synthetic-md_age-b}}{}%
\includegraphics[width=1\textwidth]{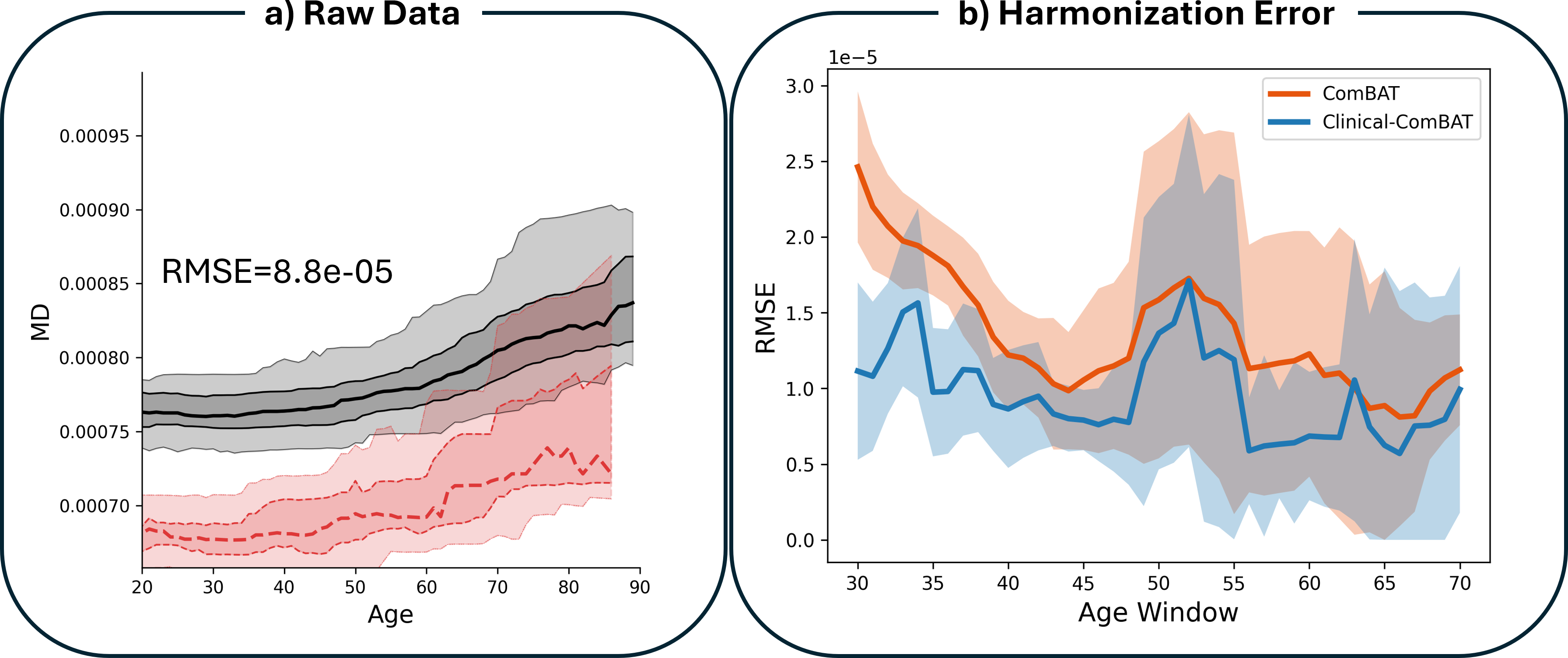}
\caption{Harmonization performance of ComBAT and Clinical-ComBAT on synthetic mean diffusivity (MD) data of the with limited age range training data. The test set comprises 100 randomly selected subjects, and the remaining 341 subjects are used to generate training sets. a) Target site data (black) and raw moving site data (red) of the white matter skeleton mask under simulated biases: slope $S = 1$, multiplicative $M = 1.50$, and additive $A = 0.90$. b) Root Mean Squared Error (RMSE) when training subjects are restricted to a $\pm$10-year age window. The plot show the mean RMSE for each window (center) across the 42 bundles. The shaded area represent one standard deviation. Clinical-ComBAT results are shown for $\tau=1,75$.}\label{fig:results-synthetic-md_age}
\end{figure}

In some cases, the training data may be limited to a narrow age range, with unseen data falling outside that range. It is therefore critical to use a harmonization model that generalizes well. Figure \ref{fig:results-synthetic-md_age} shows the harmonization performance of ComBAT and Clinical-ComBAT on synthetic MD data when training is restricted to a limited age range. The test set includes 100 randomly selected subjects and the remaining 341 subjects are use for generation training sets. Figure \ref{fig:results-synthetic-md_age-a}) displays the target site data (black) and raw moving site data (red) of the white matter skeleton mask under simulated biases ($S = 1$, $M = 1.50$, $A = 0.90$). Figure \ref{fig:results-synthetic-md_age-b}) reports RMSE across training sets restricted to a $\pm$10-year age window, plotted as a function of the window center. The lines show the mean across all 42 bundles, and the shaded area represent one standard deviation. Clinical-ComBAT results are shown for $\tau=1.75$. 
Clinical-ComBAT uses the reference slope as a prior to provide a more reliable model outside the training age window, outperforming ComBAT.

\subsection{Effect of the Bayesian Priors}

\subsubsection{Variance prior}
\label{ssec:nu}
\begin{figure}[thp]
\centering
\captionsetup[subfigure]{labelformat=empty}%
\subcaptionbox{\label{fig:results-nu-a}}{}%
\subcaptionbox{\label{fig:results-nu-b}}{}%
\subcaptionbox{\label{fig:results-nu-c}}{}%
\subcaptionbox{\label{fig:results-nu-d}}{}%
\subcaptionbox{\label{fig:results-nu-e}}{}%
\subcaptionbox{\label{fig:results-nu-f}}{}%
\includegraphics[width=1\textwidth]{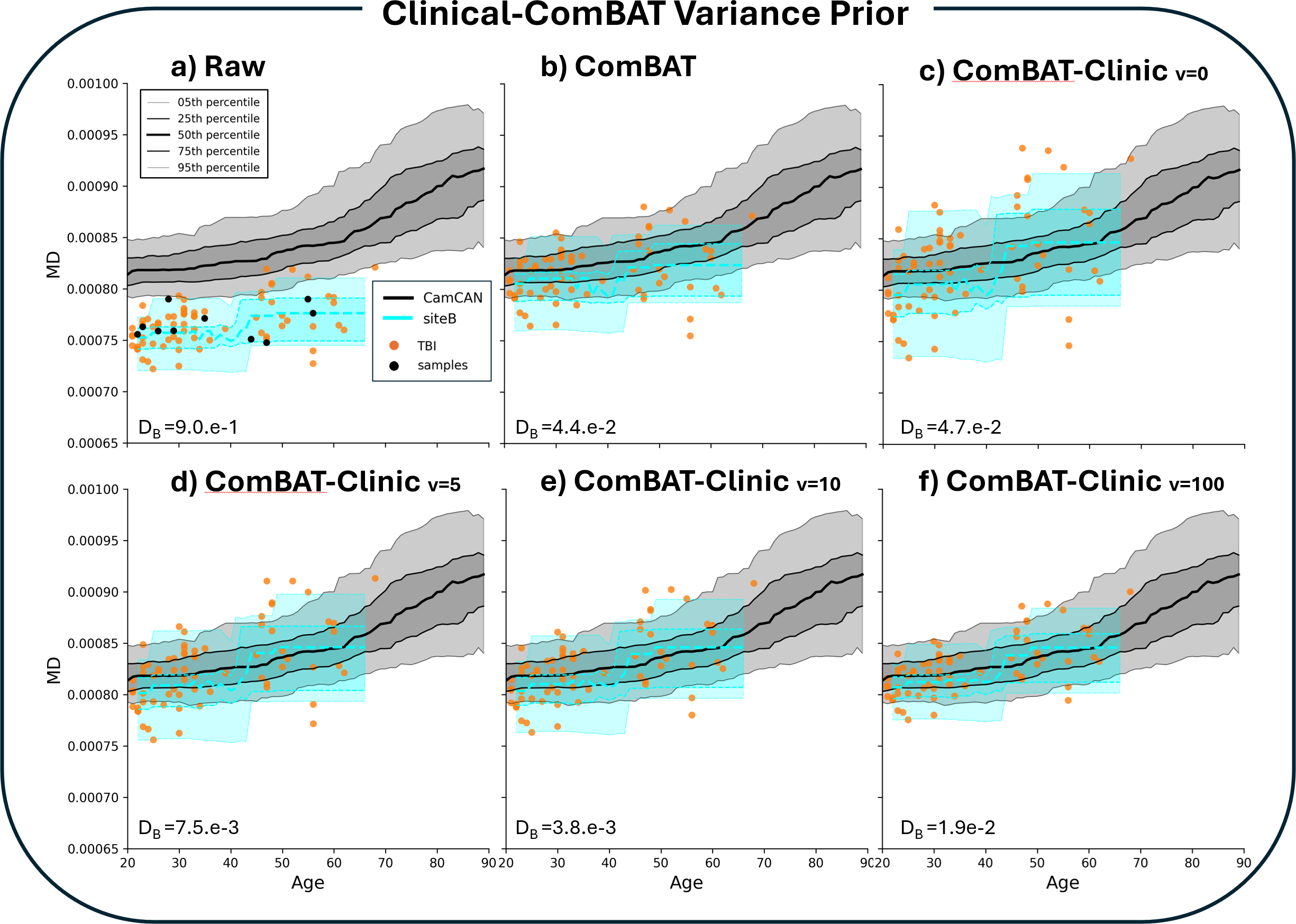}
\caption{Harmonization with Clinical-ComBAT of a limited number of ten healthy control subjects from the Track TBI dataset (site B) on the mean diffusivity (MD) of the right Corticospinal Tract (CST) bundle. (a) show the non-harmonized data, with ten randomly selected HC used for harmonization highlighted with the green circle markers. The harmonization with ComBAT is shown in (b). (c-f) show the harmonization with Clinical-ComBAT using increasing $\nu$ values (c.f Eq.~\ref{eq:nu}). With $\nu=0$, the standard deviation of the moving site is overestimated and overcorrected. With $\nu=5$ and $\nu=10$, results show the standard deviation of the moving site aligned with the standard deviation of the reference site. (f) shows an underestimation of the standard deviation due to an excessively strong a priori ($\nu=100$).}\label{fig:results-nu}
\end{figure}

Figure \ref{fig:results-nu} illustrates the effect of Clinical-ComBAT’s variance prior ($\nu$) on mean diffusivity (MD) in the right corticospinal tract (CST) bundle using a limited number of HCs. We selected ten randomly sampled HCs from the Track-TBI dataset (site B) and tested different values of the variance prior parameter (see Eq. \ref{eq:nu}). This parameter controls the weight of the prior on the variance estimation of the moving site ($d_{Mv}^2$). ComBAT (Figure \ref{fig:results-nu-b}) estimates variance using empirical Bayes, requiring simultaneous harmonization of all regions or voxels. In contrast, Clinical-ComBAT leverages the well-populated reference site to estimate the variance while accounting for the number of available subjects in the moving site, thereby enabling independent harmonization of regions or voxels. Furthermore, it supports single-region harmonization without compromising performance.

In Figure \ref{fig:results-nu-c}, setting $\nu=0$ underestimates $d_{Mv}^2$, leading to an insufficient correction of the multiplicative bias, as indicated by the wider percentile lines compared to the reference site. With $\nu=5$ and $\nu=10$ (Figures \ref{fig:results-nu-d} and \ref{fig:results-nu-e}), Clinical-ComBAT aligns $d_{Mv}^2$ with $d_{Rv}^2$. However, at $\nu=100$ (Figure \ref{fig:results-nu-f}) $d_{Mv}^2$ is overestimated due to excessive reliance on the reference prior. A high $\nu$ with few moving site subjects effectively fixes $d_{Mv}^2$ to $d_{Rv}^2$, preventing proper correction of the multiplicative bias.

\subsubsection{Covariate Prior}
\label{sec:tau}
\begin{figure}[!tbh]
\centering
\includegraphics[width=1\textwidth]{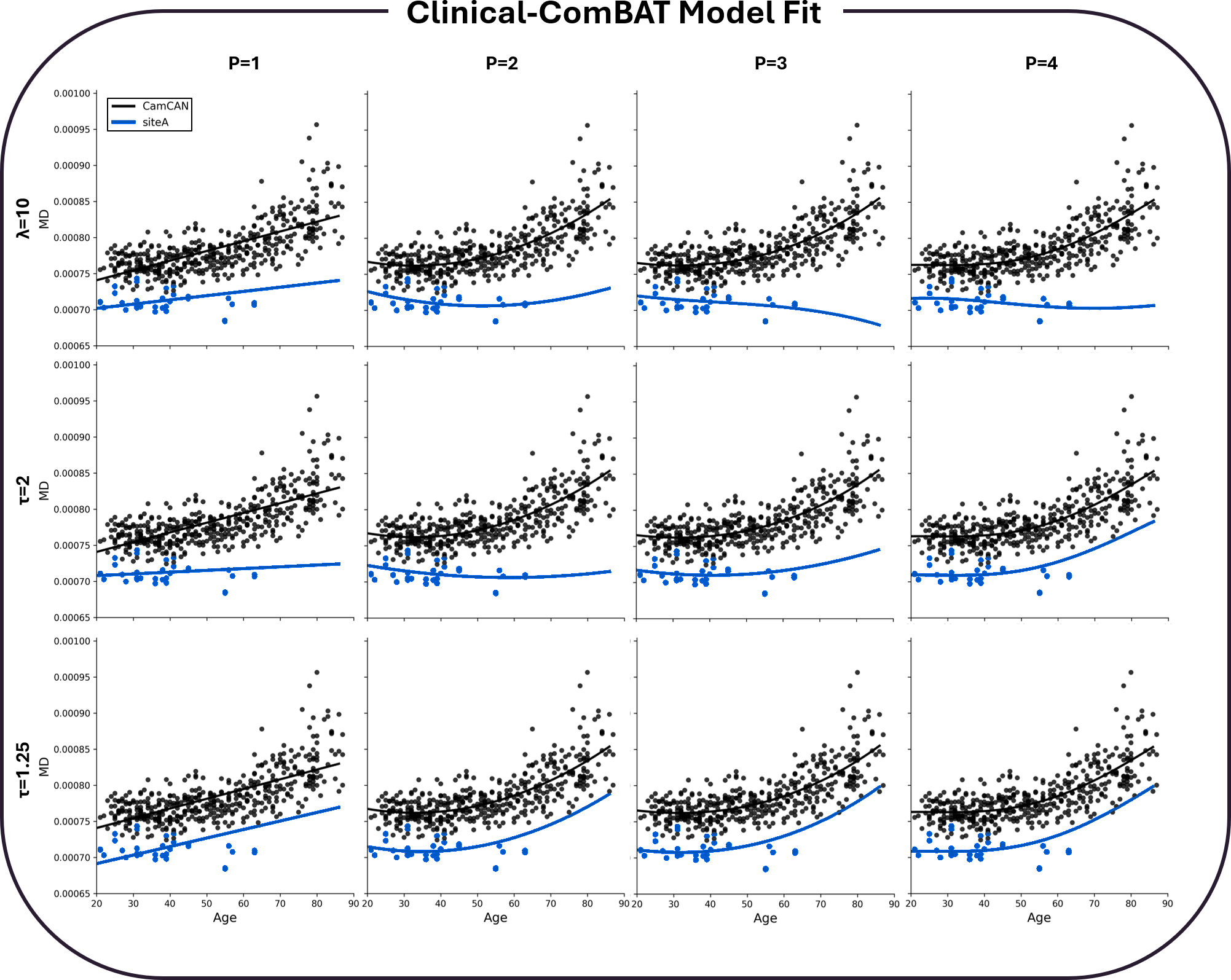}
\caption{Clinical-ComBAT model fit for the mean diffusivity (MD) of the white matter skeleton mask of Track TBI site A, using reference site a priori with and without automatic hyperparameter tuning with polynomial order P=1 to 4. The first row shows the fitted model using a fixed $\lambda=10$. The second and third rows show the fitted model using automatic hyperparameter tuning ($\tau=2$, and $\tau=1.25$, respectively)}\label{fig:results-order}
\end{figure}

Figure \ref{fig:results-order} illustrates the importance of the Clinical-ComBAT hyperparameter auto-tuning when dealing with a moving site whose HC subjects spans a limited age range (21-63 years old). It shows the effects of various $\lambda$ values with a fixed $\nu=5$ using polynomial models of order $P=1$ to $P=4$. Although a fixed $\lambda=10$ shows a good model fit (green line) within the moving site's age range, the model quickly diverges as the age deviates from the input data. A polynomial degree of $P=2$ or $P=3$ with auto-tuning $\tau=2$ provides a good balance, closely fitting the available data while yielding reasonable extrapolations. Higher polynomial degrees ($P=3$ and $P=4$) require auto-tuning for usability beyond the input data range. While higher $\tau$ values improve extrapolation, they reduce fit quality within the data range due to the stronger prior. Overall, data models estimated with $\tau=2$ and $\tau=1.25$ are suitable for the harmonization of moving site subjects of all ages.

\section{Discussion}

Clinical-ComBAT improves harmonization performance across a range of bias scenarios and data limitations compared to ComBAT \cite{Fortin2017}. In ComBAT, the regression parameter $\beta_v$ in is identical across all sites. Clinical-ComBAT introduces a subtle but crucial modification by adding a site index $i$, resulting in $\beta_{iv}$. This adjustment allows each site to have its own regression parameter, which is essential for addressing site-specific variations. This site-specific regression parameters allow Clinical-ComBAT to correct slope-related biases that ComBAT fails to address (Figs. \ref{fig:results-md} and \ref{fig:results-synthetic-md}). Moreover, Clinical-ComBAT shows superior generalization to unseen data, outperforming ComBAT with limited sample sizes (Figs. \ref{fig:results-synthetic-md_number} and \ref{fig:nbr-sub}) and restricted age ranges scenarios (Fig. \ref{fig:results-synthetic-md_age}), common in clinical datasets. This robustness is enabled by the integration of priors from the reference site, which stabilize parameter estimation in challenging settings. Furthermore, the adaptive regularization of variance ($\nu$) and regression coefficients ($\tau$) provides a flexible framework that supports accurate extrapolation (Figs. \ref{fig:results-nu} and \ref{fig:results-order}). Together, these features position Clinical-ComBAT as a reliable harmonization approach for normative modeling applications in heterogeneous multi-site clinical cohorts. Moreover, the improved alignment provided by Clinical-ComBAT may facilitate the delineation of pathological subjects from the reference population. 

In this study, we used the Cambridge Centre for Ageing and Neuroscience (CamCAN) dataset \cite{Shafto2014, Taylor2017} as the reference site to demonstrate the potential of Clinical-ComBAT. This dataset was chosen primarily for its large sample of healthy controls spanning a wide age range (18–87 years). In future work, other datasets, particularly those providing richer DW-MRI acquisitions, could be used to enable harmonization of additional diffusion metrics. More broadly, the DW-MRI community may eventually converge on a standard reference dataset, facilitating direct comparability of reported values across studies, thereby improving reproducibility and reducing site-related biases.

\subsection{Limitations and Recommendations}
Clinical-ComBAT leverages prior information from a well-characterized reference site to stabilize parameter estimation when moving site data are sparse or incomplete. Low values of the hyperparameter $\tau$ constrain the regression coefficients to those of the reference site, effectively fixing the model shape and estimating only the intercept. Similarly, high values of $\nu$ fix the moving site variance to that of the reference site. While these priors enable extrapolation and prevent overfitting, they may introduce bias if the true distribution of the moving site differs substantially in variance or covariate effects. Nonetheless, we believe extrapolating from the reference site is preferable to relying on poorly estimated parameters, especially in underpowered scenarios. The harmonization model should be retrained as additional moving site data become available, allowing for refinement of site-specific parameters and improved accuracy.

In practice, we recommend selecting the polynomial degree for modeling based on the reference site, choosing the minimal order that ensures a good fit (e.g., based on Bhattacharyya distance). For mean diffusivity (MD), fractional anisotropy (FA), and apparent fiber density (AFD), a second-order polynomial ($P=2$) is typically sufficient. Regarding variance estimation, our results suggest setting $\nu = 5$, which allows $d_{Mv}^2$ to be primarily data-driven when sufficient observations are present, while gradually incorporating the reference prior in low-data scenarios. For the regression prior, a value of $\tau = 2$ offers a robust balance between model flexibility and extrapolation. Lower $\tau$ values risk over-constraining the model, while higher values may produce unstable predictions.

\begin{table}[tp]
\centering
\caption{The nine (9) differences between Clinical-ComBAT and ComBAT.}
\footnotesize
\label{tab:contrib}
\begin{tabular}{|p{1.9cm}||p{5.3cm}|p{5.3cm}|}
 \hline
 & \textbf{Clinical-ComBAT} & \textbf{ComBAT} \\ \hline
Type of harmonization & Pairwise harmonization between a moving site and a well-populated reference site & Group harmonization between a set of two or more sites \\ \hline
Data formation model & New data formation model, Eq.~(\ref{eq:data_formation}), with one additive and one multiplicative bias & Data formation model of Eq.~(\ref{eq:combat}) with two additive and two multiplicative biases \\ \hline
Model parameters & Model parameter vector $\vec{\beta}_{iv}$ is site specific & Model parameter vector $\vec{\beta}_{v}$ is shared across all sites \\ \hline
Model type & Non-linear model, Eq.~(\ref{eq:phi}) & Linear model, Eq.~(\ref{eq:combat2}) \\ \hline
Variance of reference site & Variance $d_{Rv}^2$ computed via maximum likelihood, Eq.~(\ref{eq:varrefsite}) & Variance of each site computed via maximum a posteriori with inverse-Gamma prior, Eq.~(S.18) \\ \hline
Parameter estimation & Moving site parameters $\vec{\beta}_{Mv}$ computed via MAP using moving-site data and a prior on reference-site parameters $\vec{\beta}_{Rv}$ (Eq.~(\ref{eq:betaMv})) & Parameters $\vec{\beta}_{v}$ computed via ordinary maximum likelihood pooling data from all sites (Eq.~(S.1)) \\ \hline
Variance of moving site & Variance $d_{Mv}^2$ computed via MAP using a Gamma prior conditioned on $d_{Rv}^2$ (Eq.~(\ref{eq:nu})) & Variance of each site computed via maximum a posteriori with inverse-Gamma prior, Eq.~(S.18) \\ \hline
 \begin{tabular}{l}
    Quality\\control
  \end{tabular}
 & Includes a QC module evaluating goodness of fit between target and harmonized moving site (Algo.~3) & No quality control module \\ \hline
Hyperpara-meter tuning & Some hyperparameters (e.g., $\vec{\lambda}$ and $\nu_0$) automatically estimated via auto-tuning algorithm (Algo.~4) & No hyperparameter auto-tuning \\ \hline
\end{tabular}
\end{table}

\section{Conclusion}\label{sec:conclusion}
This study introduces Clinical-ComBAT, an adaptation of the widely used ComBAT method designed to address key limitations in diffusion MRI harmonization for clinical applications.  As summarized in Table~\ref{tab:contrib}, Clinical-ComBAT has nine key differences with the original ComBat method.

While ComBAT effectively reduces site-related biases, its assumptions, such as linear covariate relationships, large sample requirements, and fixed site numbers, limit its applicability in real-world clinical settings. Clinical-ComBAT overcomes these constraints by incorporating a polynomial data model and a new mathematical formulation for estimating the harmonization parameters taylored to site-specific harmonization to a normative reference, and a variance prior tailored for small sample sizes.  Unlike ComBAT, which harmonizes all bundles (or voxels, or regions) simultaneously, Clinical-ComBAT processes each independently. This independence allows for increasing the number of bundles without altering harmonization parameters for existing ones, and enables single-bundle harmonization without degrading performance. Our results demonstrate that Clinical-ComBAT improves harmonization quality, particularly in datasets with limited healthy controls, reducing residual biases and better aligning diffusion metrics with the reference distribution. These findings highlight the method’s potential to enhance cross-site comparability in clinical studies, ultimately improving the detection and interpretation of disease-related abnormalities.

\section*{Data and Code Availability}
All MRI dataset used in this work are publicly available. The source code and documentation of the Clinical-ComBAT method is available at \url{github.com/scil-vital/clinical-ComBAT/}.

\section*{Author Contributions}
GG contributed to conceptualization, methodology, software development, and writing of the original draft. ME contributed to methodology, software development, and visualization. FD contributed to conceptualization, methodology, and software development. MD and GT contributed to data curation. YD contributed to software development. JCH contributed to supervision. AB contributed to software development and data curation. MD contributed to conceptualization, supervision, and funding acquisition. PMJ contributed to conceptualization, methodology, writing of the original draft, supervision, and funding acquisition.
\section*{Acknowledgement}
This research was partially funded by the {\it Consortium québécois sur la découverte du médicament} (CQDM) and the Canadian NSERC discovery grants 	RGPIN-2023-04584 and RGPIN-2020-04818. This research was also enabled by support provided by the Digital Research Alliance of Canada.

\section*{Declaration of Competing Interests}
Matthieu Dumont, Guillaume Theaud, Jean-Christophe Houde, Maxime Descoteaux and Pierre-Marc Jodoin report membership and employment with Imeka Solutions inc.

Gabriel Girard, Manon Edde, Félix Dumais, Matthieu Dumont, Guillaume Theaud, Yoan David, Jean-Christophe Houde, Maxime Descoteaux and Pierre-Marc Jodoin have a patent pending for the Clinical-ComBAT method: (2024) Harmonization of diffusion MRI, USPTO Serial No.63/709,998.

\bibliographystyle{ieeetr} 
\bibliography{ref}
\newpage

\section*{Supplementary Material}

\section{Location and Scale (L/S) ComBAT}
\label{ssec:lscombat}
Follow up of section 2.1 in the paper, L/S ComBAT (c.f. section 2.3 of \cite{Johnson2007})  estimates the parameters $\hat\alpha_v$,$\hat\beta_v$, $\hat\gamma_{iv}$ and $\hat\delta_{iv}$ following a maximum likelihood of Gaussian distributions. As such, the regression vector $\hat\beta_v$ and the intercept $\hat\alpha_v$ are obtained with an ordinary least-square approach. This involves pooling data from every location $v$ (voxel or brain region) and from every patient $j$ across all sites $i$, and use the usual regression solution \cite{Bishop2006}:
\begin{equation}
\label{seq:lsab}
    [\hat\alpha_v, \hat\beta_v]^T = (X^TX)^{-1}X^Ty_v,
\end{equation}
where $X$ is the matrix of covariate vectors from every participant across all sites, and $y_v$ is a vector containing the metric associated with location $v$ for every participant in all sites. According to this framework, the global intercept $\hat\alpha_v$ represents the average feature value at location $v$ for all images across all sites, excluding the effect of covariates, namely
\begin{equation}
\label{seq:lsa}
    \hat\alpha_v = \frac{1}{J_I}\sum_{ij}(y_{ijv} - x^T_{ij}\hat\beta_v),
\end{equation}
where $J_I$ is the total number of subjects across all sites.

The additive and multiplicative site-specific biases $\hat\gamma_{iv}$ and $\hat\delta_{iv}$ are determined by taking the mean and variance of the rectified value at location $v$ for all images acquired at site $i$:
\begin{eqnarray}
    \label{seq:lsgamma}
    \hat\gamma_{iv} = \frac{1}{J_i}\sum_j(y_{ijv}-\hat\alpha_v-x^T_{ij}\hat\beta_v), \\
    \label{seq:lsdelta}
    \hat\delta^2_{iv} = \frac{1}{J_i -1} \sum_i (y_{ijv} - \hat\alpha_v - x^T_{ij}\hat\beta_v - \hat\gamma_{iv})^2.
\end{eqnarray}

Once these parameters estimated (namely $\hat{\alpha}_v$, $\hat{\beta_v}$, $\hat{\gamma}_{iv}$, and $\hat{\delta}_{iv}$), the harmonized values $y_{ijv}^{harm}$ can be computed as follows (Eq(3) in the paper):
\begin{equation}
\label{eq:lscombat}
    y^{harm}_{ijv} = \frac{y_{ijv}-\hat{\alpha}_v - \vec x^T_{ij} \hat{\beta}_v - \hat{\gamma}_{iv}}{\hat{\delta}_{iv}} + \hat{\alpha}_v + \vec x^T_{ij}  \hat{\beta}_v.
\end{equation}

The reader shall note that when the total number of images $J_I$ acquired across all sites is sufficiently large, the maximum likelihood estimate of $\hat\beta_v$ and $\hat\alpha_v$ from Eqs \ref{seq:lsab} and \ref{seq:lsa} can be considered unbiased. However, for any site $i$ with a low number of acquired images (a common scenario in clinical practice) $\hat\gamma_{iv}$ and $\hat\delta_{iv}$ are biased and thus should not be relied upon.
One common solution to that problem is to estimate $\hat\gamma_{iv}$ and $\hat\delta_{iv}$ following a Bayesian maximum a posteriori formulation while keeping as is Eq(\ref{seq:lsab}) and (\ref{seq:lsa}) for estimating $\hat\beta_v$ and $\hat\alpha_v$ \cite{Fortin2017,Johnson2007,Jodoin2025}.  Details on the Bayesian formulation are provided in section S.2.

\section{ComBAT}
\label{sec:vanillacombat}
Follow up of section 2.1 in the paper.  For a step-by-step explanation, please refer to figure 1 of ~\cite{Jodoin2025}. This section explains the Bayesian approach of ComBAT for estimating the harmonization parameters $\hat{\alpha}_v$, $\hat{\beta_v}$, $\hat{\gamma}_{iv}$, $\hat{\sigma}_{v}$, and $\hat{\delta}_{iv}$.
To do so, ComBAT begins by standardizing the data, which involves subtracting the population intercept, regression weight and variance, represented mathematically as
\begin{equation}
    z_{ijv} = \frac{y_{ijv}-\hat\alpha_v-x^T_{ij}\hat\beta_v}{\hat\sigma_v},
\end{equation}
where the variance is determined with the usual empitical equation: 
\begin{equation}
    \hat\sigma^2_v = \frac{1}{J_I}\sum_{ij}(y_{ijv} - \hat\alpha_v - x^T_{ij}\hat\beta_v - \hat\gamma_{iv})^2,
\end{equation}
and $\hat\gamma_{iv}$ is calculated following Eq~(\ref{seq:lsgamma}).

At this stage, the bias of each site 
is denoted as $\gamma^*_{iv}$ and the variance of each site is $\delta^{2*}_{iv}$. These two variables are estimated by maximizing their posterior distributions rather than their likelihood distributions. According to Bayes’ theorem, the posteriors of $\gamma^*_{iv}$ and $\delta^{2*}_{iv}$ can be expressed as
\begin{equation}
    \label{seq:gammastar}
    P(\gamma^*_{iv} | z_{iv}, \delta^{2*}_{iv}) \propto P(z_{iv} | \gamma^*_{iv}, \delta^{2*}_{iv}) P(\gamma^*_{iv}),
\end{equation}
\begin{equation}
    \label{seq:deltastar}
    P(\delta^{2*}_{iv} | z_{iv}, \gamma^*_{iv}) \propto P(z_{iv} | \gamma^*_{iv}, \delta^{2*}_{iv}) P(\delta^{2*}_{iv}),
\end{equation}
where
\begin{equation}
    P(z_{iv}| \gamma^*_{iv}, \delta^{2*}_{iv}) = \mathcal{N}(\gamma^*_{iv}, \delta^{2*}_{iv}),
\end{equation}
\begin{equation}
    P(\gamma^*_{iv}) = \mathcal{N}(\mu_i, \tau_i^2),
\end{equation}
\begin{equation}
    P(\delta^{2*}_{iv}) = \mathcal{IG}(\lambda_i, \theta_i),
\end{equation}
are a Gaussian likelihood, a Gaussian prior and an inverse Gamma prior, respectively.

The hyperparameters of the prior distributions $\mu_i$, $\tau_i^2$, $\lambda_i$, and $\theta_i$ are estimated based on the moment of these distributions. Specifically, $\mu_i$ and $\tau_i^2$ are derived from the first and second moment of a Gaussian distribution, namely
\begin{equation}
    \label{seq:mutau}
    \bar\mu_i = \frac{1}{v}\sum_v \hat\gamma^*_{iv} \;\;\;\;\;\; \bar\tau^2_i = \frac{1}{V-1}\sum_v (\hat\gamma^*_{iv} - \bar\mu_i)^2
\end{equation}

where $V$ is the number of brain regions (or voxels) and $\hat\gamma^*_{iv} = \frac{1}{J_i}\sum_j z_{ijv}$ is the region-wise and sitewise model intercept of the standardized data and should not be confused with $\hat\gamma_{iv}$ from Eq~(\ref{seq:lsgamma}).

To estimate $\lambda_i$ and $\theta_i$, one must first calculate the voxel-wise and sitewise standardized variance $\delta^{2*}_{iv} = \frac{1}{J_i - 1}\sum_j(z_{ijv}-\hat\gamma^*_{iv})^2$. This variance shall not be confused with $\hat \delta^{2}_{iv}$ from Eq~( \ref{seq:lsdelta}). The empirical mean and variance of $\delta^{2*}_{iv}$ across all voxels is computed as $\bar{G}_I = \frac{1}{V}\sum_v\delta^{2*}_{iv}$ and $\bar{S}^2_I = \frac{1}{V-1}\sum_v (\delta^{2*}_{iv} - \bar{G}_I)^2$. Then, by equating $\bar{G}_I$ and $\bar{V}^2_i$ to the first and second theoretical moments of the inverse gamma distribution, we get
\begin{equation}
    \label{seq:GS}
    \bar{G}_i = \frac{\theta_i}{\lambda_i - 1} \;\;\;\;\;\; \bar{S}_i^2 = \frac{\theta^2_i}{(\lambda_i - 1)^2(\lambda_i - 2)},
\end{equation}
which can be rearranged to estimate the two hyperparameters,
\begin{equation}
    \bar\lambda_i = \frac{\bar{G}_i^2 + 2\bar{S}^2_i}{\bar{S}^2_i}
\end{equation}
\begin{equation}
    \bar\theta_i = \frac{\bar{G}^3_i + \bar{GS}^2_i}{\bar{S}_i^2}.
\end{equation}

Now that the hyperparameters of the likelihood and prior distributions have been estimated, by expanding Eqs~(\ref{seq:gammastar}) and (\ref{seq:deltastar}) and integrating them with Eqs~(\ref{seq:mutau}) and (\ref{seq:GS}), we arrive at the mathematical expectation of the posterior distributions, which yields the estimates of $\gamma^*_{iv}$ and $\delta^{2*}_{iv}$:
\begin{equation}
    \label{seq:gammabar}
    \bar\gamma^*_{iv} = \mathbb{\hat{E}}(\gamma^*_{iv} | z_{iv}, \delta^{2*}_{iv}) = \frac{J_i\hat\tau^2_i\hat\gamma^*_{iv} + \bar\delta^{2*}_{iv}\bar\mu_i}{J_i\bar\tau^2_i + \bar\delta^{2*}_{iv}}
\end{equation}
\begin{equation}
    \label{seq:deltabar}
    \bar\delta^{2*}_{iv} = \mathbb{\hat{E}}(\delta^{2*}_{iv} | z_{ijv}, \gamma^*_{iv}) = \frac{\bar\theta_i + \frac{1}{2}\sum_j(z_{ijv} - \bar\gamma^*_{iv})^2}{\frac{J_i}{2} +\bar\lambda_i - 1}.
\end{equation}

Given the interdependency of Eqs~(\ref{seq:gammabar}) and (\ref{seq:deltabar}), $\bar\gamma^*_{iv}$ and $\bar\delta^{2*}_{iv}$ are computed through an iterative process. This starts with a reasonable initial value for $\bar\delta^{2*}_{iv}$ (e.g., $\hat\delta^{2*}_{iv}$), followed by the calculation of $\gamma^*_{iv}$, and then re-estimation of $\bar\delta^{2*}_{iv}$ using the new $\gamma^*_{iv}$ value, and so forth. This cycle is repeated until convergence is achieved.

Once all ComBAT parameters  have been empirically estimated, data harmonization is performed as follows:
\begin{equation}
    y^{ComBAT}_{ijv} = \frac{\hat\sigma_v}{\bar\delta^*_{iv}} (z_{ijv} - \bar\gamma^*_{iv}) + \hat\sigma_v + x^T_{ij}\hat\beta_v.
\end{equation}

The reader shall notice that this harmonization function is different from that of L/S Combat in Eq.~(\ref{eq:lscombat}).

\section{Maximum a posteriori for the estimation of the moving site parameter $\vec{\beta}_{Mv}$}

As mentioned in Section~2.3, $\vec{\beta}_{Mv}$ is estimated by maximizing the posterior distribution 
$P(\vec{\beta}_{Mv}\,|\,D_{Mv})$ where 
$D_{Mv} = \{ Y_{Mv}, X_M \}$ contains the list of all values and covariables of the moving site $M$ at location $v$. 
Following Bayes’ theorem, one can reformulate the posterior as
\begin{equation}
P(\vec{\beta}_{Mv}\,|\,D_{Mv}) = 
\frac{P(D_{Mv}\,|\,\vec{\beta}_{Mv}) \, P(\vec{\beta}_{Mv})}{P(D_{Mv})}, 
\label{eq:a1}
\end{equation}
where $P(D_{Mv}\,|\,\vec{\beta}_{Mv})$ is the likelihood pdf, $P(\vec{\beta}_{Mv})$ is the prior pdf, and $P(D_{Mv})$ is the evidence. 
Since $P(D_{Mv})$ is independent from the to-be-optimized parameters $\vec{\beta}_{Mv}$, it can be considered a multiplicative factor and thus
\begin{equation}
P(\vec{\beta}_{Mv}\,|\,D_{Mv}) \propto P(D_{Mv}\,|\,\vec{\beta}_{Mv}) \, P(\vec{\beta}_{Mv}). 
\label{eq:a2}
\end{equation}

Since both $P(D_{Mv}\,|\,\vec{\beta}_{Mv})$ and $P(\vec{\beta}_{Mv})$ are Gaussian distributions (see Eqs.~(11) and (12) in the paper), the posterior is also Gaussian. Starting with the likelihood pdf of the entire population $D_{Mv}$,
\begin{equation}
P(D_{Mv}\,|\,\vec{\beta}_{Mv}) = \prod_{j=1}^{J_M} P(\vec{\phi}_{Mj}, y_{Mjv}\,|\,\vec{\beta}_{Mv}), 
\label{eq:a3}
\end{equation}
and since $P(\vec{\phi}_{Mj}, y_{Mjv}\,|\,\vec{\beta}_{Mv})$ is Gaussian for each data point,
\begin{equation}
P(D_{Mv}\,|\,\vec{\beta}_{Mv}) = \prod_{j=1}^{J_M} \frac{1}{\sqrt{2\pi}\,\sigma_M}
\exp\!\left(-\frac{(y_{Mjv} - \vec{\beta}_{Mv}^T \vec{\phi}_{Mj})^2}{2\sigma_M^2}\right), 
\label{eq:a4}
\end{equation}
which simplifies to
\begin{equation}
P(D_{Mv}\,|\,\vec{\beta}_{Mv}) = \frac{1}{(2\pi\sigma_M^2)^{J_M/2}} 
\exp\!\left(-\frac{1}{2\sigma_M^2}\sum_{j=1}^{J_M}(y_{Mjv} - \vec{\beta}_{Mv}^T \vec{\phi}_{Mj})^2\right). 
\label{eq:a5}
\end{equation}

The prior distribution is also Gaussian, with a mean vector of $\vec{\beta}_{Rv}$ (Eq.~12) and covariance $\Sigma_0$:
\begin{equation}
P(\vec{\beta}_{Mv}) = 
\frac{1}{(2\pi)^{P/2}|\Sigma_0|^{1/2}}
\exp\!\left(-\frac{1}{2}(\vec{\beta}_{Mv}-\vec{\beta}_{Rv})^T \Sigma_0^{-1}(\vec{\beta}_{Mv}-\vec{\beta}_{Rv})\right).
\label{eq:a6}
\end{equation}

Multiplying Eqs.~\eqref{eq:a5} and \eqref{eq:a6}, the posterior becomes
\begin{equation}
P(\vec{\beta}_{Mv}\,|\,D_{Mv}) \propto 
\exp\!\left(-\frac{1}{2\sigma_M^2}\sum_{j=1}^{J_M}(y_{Mjv}-\vec{\beta}_{Mv}^T\vec{\phi}_{Mj})^2 
-\frac{1}{2}(\vec{\beta}_{Mv}-\vec{\beta}_{Rv})^T\Sigma_0^{-1}(\vec{\beta}_{Mv}-\vec{\beta}_{Rv})\right).
\label{eq:a8}
\end{equation}

Maximizing Eq.~\eqref{eq:a8} is equivalent to minimizing its negative log:
\begin{equation}
\vec{\beta}_{Mv} = \arg\min_{\vec{\beta}_{Mv}}
\frac{1}{2\sigma_M^2}\sum_{j=1}^{J_M}(y_{Mjv}-\vec{\beta}_{Mv}^T\vec{\phi}_{Mj})^2
+\frac{1}{2}(\vec{\beta}_{Mv}-\vec{\beta}_{Rv})^T\Sigma_0^{-1}(\vec{\beta}_{Mv}-\vec{\beta}_{Rv})
\label{eq:a9}
\end{equation}
and assuming $\Sigma_0$ is diagonal, the equation reduces to
\begin{equation}
\vec{\beta}_{Mv} = \arg\min_{\vec{\beta}_{Mv}} 
\Bigg[ \sum_{j=1}^{J_M}(y_{Mjv}-\vec{\beta}_{Mv}^T\vec{\phi}_{Mj})^2 
+ \vec{\lambda}^T (\vec{\beta}_{Mv}-\vec{\beta}_{Rv})^T(\vec{\beta}_{Mv}-\vec{\beta}_{Rv}) \Bigg],
\label{eq:a10}
\end{equation}
with $\vec{\lambda}\in\mathbb{R}^{P+}$ a vector of hyperparameters.  
By forcing $\nabla_{\vec{\beta}_{Mv}} L(\vec{\beta}_{Mv})=0$, one arrives at
\begin{equation}
\vec{\beta}_{Mv}^T = (\Phi_M^T \Phi_M + \vec{\lambda}^T I)^{-1}(\Phi_M^T \vec{Y}_{Mv} + \vec{\lambda}^T \vec{\beta}_{Rv}), 
\label{eq:a15}
\end{equation}
which corresponds to Eq.~(13) in the paper.

\section*{Maximum a posteriori of the moving site variance parameter $d_{Mv}^2$}

 As mentioned in Section~2.3, the moving site variance $d_{Mv}^2$ is estimated by maximizing the posterior distribution $P(d_{Mv}^2\,|\,Z_{Mv})$, where 
$Z_{Mv} = \{ z_{Mjv} = y_{Mjv} - \vec{\phi}_{Mj}^T \vec{\beta}_{Mv},\;\; \forall j\}$.
Following Bayes’ theorem:
\begin{equation}
P(d_{Mv}^2\,|\,Z_{Mv}) = \frac{P(Z_{Mv}\,|\,d_{Mv}^2) P(d_{Mv}^2)}{P(Z_{Mv})}.
\label{eq:a16}
\end{equation}

Since $Z_{Mv}$ is a zero-centered Gaussian with variance $d_{Mv}^2$, we assume that
\begin{equation}
P(Z_{Mv}\,|\,d_{Mv}^2) = \prod_{j=1}^{J_M} \frac{1}{\sqrt{2\pi} d_{Mv}}
\exp\!\left(-\frac{z_{Mjv}^2}{2d_{Mv}^2}\right).
\label{eq:a20}
\end{equation}

If we use a Gamma function as the prior $P(d_{Mv}^2)$ with $\tau = 1/d_{Mv}^2$, we get that the  prior of the variance of the moving site is driven by the unbiased estimation of the variance of the reference site computer by Eq.(9).  This leads to :
\begin{equation}
P(\tau) = \frac{b_0^{a_0}}{\Gamma(a_0)} \tau^{a_0-1} \exp(-b_0\tau).
\label{eq:a19}
\end{equation}

Since the posterior is the product of a Gaussian and a Gamma distribution, it remains Gamma-distributed with parameters
\[
a_n = a_0 + \frac{J_M}{2}, \quad b_n = b_0 + \frac{1}{2}\sum_{j=1}^{J_M} z_{Mjv}^2.
\]

Thus the MAP estimate of $\tau$ is
\begin{equation}
\tau = \frac{1}{d^2_{Mv}} =\frac{a_0 + J_M/2}{b_0 + \tfrac{1}{2}\underbrace{\Sigma_{j=1}^{J_M} z_{Mjv}^2}_{J_M \hat d^2_{Mv}}}, 
\label{eq:a24}
\end{equation}
leading to
\begin{equation}
d_{Mv}^2 = \frac{2b_0 + J_M \hat{d}_{Mv}^2}{2a_0 + J_M}, 
\label{eq:a25}
\end{equation}
and, by setting $a_0 = \nu_0/2$ and $b_0 = d_{Tv}^2\nu_0/2$, the last equation simplifies to
\begin{equation}
d_{Mv}^2 = \frac{\nu_0 d_{Tv}^2}{\nu_0 + J_M} + \frac{J_M \hat{d}_{Mv}^2}{\nu_0 + J_M},
\label{eq:a26}
\end{equation}
which corresponds to Eq.~(17) in the paper.

\section{Clinical-ComBAT algorithms}
\begin{algorithm}[H]
\caption{Clinical-ComBAT-Fit}

\KwIn{}
$D_{Tv} = \{ (y_{T1v}, \vec{x}_{T1}), \dots, (y_{TN_Tv}, \vec{x}_{TJ_T}) \}$ \tcp*{\small Target site data}
$D_{Mv} = \{ (y_{M1v}, \vec{x}_{M1}), \dots, (y_{MN_Mv}, \vec{x}_{MJ_M}) \}$ \tcp*{\small Moving site data}
$P, \vec{\lambda}, \nu_0$ \tcp*{\small Hyperparameters}

\KwOut{}
$\{ d_{Mv}^2, \vec{\beta}_{Mv}, d_{Tv}^2, \vec{\beta}_{Tv} \}$ \tcp*{\small Harmonization parameters}

\hrulefill

\textbf{Target site parameters:}\\
$\vec{\phi}_{Tj} \leftarrow (\vec{x}_{Tj}^T \cdot \vec{x}_{Tj} + 1)^P \quad \forall j$ \hfill Eq.~(19) \\
$\Phi_T \leftarrow$ stack all vectors $\vec{\phi}_{Tj}$ into a matrix \\
$\vec{\beta}_{Tv} \leftarrow (\Phi_T^T \Phi_T)^{-1} \Phi_T^T Y_{Tv}$ \hfill Eq.~(23) \\
$d_{Tv}^2 \leftarrow \frac{1}{J_T} \sum_{j=1}^{J_T} (y_{Tjv} - \vec{\phi}_{Tj}^T \vec{\beta}_{Tv})^2$ \hfill Eq.~(24) \\
\hspace{0.3cm}

\textbf{Moving site parameters:}\\
$\vec{\phi}_{Mj} \leftarrow (\vec{x}_{Mj}^T \cdot \vec{x}_{Mj} + 1)^P \quad \forall j$ \hfill Eq.~(19) \\
$\Phi_M \leftarrow$ stack all vectors $\vec{\phi}_{Mj}$ into a matrix \\
$\vec{\beta}_{Mv} \leftarrow (\Phi_M^T \Phi_M + \vec{\lambda}^T I)^{-1} (\Phi_M^T Y_{Mv} + \vec{\lambda}^T \vec{\beta}_{Tv})$ \hfill Eq.~(28) \\
$\hat{d}_{Mv}^2 \leftarrow \frac{1}{J_M} \sum_{j=1}^{J_M} (y_{Mjv} - \vec{\phi}_{Mj}^T \vec{\beta}_{Mv})^2$ \\
$d_{Mv}^2 \leftarrow \frac{J_M \hat{d}_{Mv}^2}{J_M + \nu_0} + \frac{\nu_0 d_{Tv}^2}{J_M + \nu_0}$ \hfill Eq.~(32) \\
\hspace{0.3cm}

\KwRet $\{ d_{Mv}^2, \vec{\beta}_{Mv}, d_{Tv}^2, \vec{\beta}_{Tv} \}$
\end{algorithm}

\vspace{1cm}

\begin{algorithm}[H]
\caption{Clinical-ComBAT-Apply}
\KwIn{}
    $D_{Mv} = \{ (y_{M1v}, \vec{x}_{M1}), \dots, (y_{MN_Mv}, \vec{x}_{MJ_M'}) \}$ \tcp*{\small Moving site data}
    $P$ \tcp*{\small Hyperparameter}
    $\{ d_{Mv}^2, \vec{\beta}_{Mv}, d_{Tv}^2, \vec{\beta}_{Tv} \}$  \tcp*{\small Precomputed parameters}

\KwOut{}
    $\{ \hat{y}_{M1v}, \dots, \hat{y}_{MJ_Mv} \}$ \tcp*{\small Harmonized values}

\hrulefill

\For{$j = 1 \dots J_M'$}{
    $\vec{\phi}_{Mj} \leftarrow (\vec{x}_{Mj}^T \cdot \vec{x}_{Mj} + 1)^P$ \hfill Eq.~(19) \\

    $\hat{y}_{Mjv} \leftarrow 
    \frac{(y_{Mjv} - \vec{\beta}_{Mv}^T \vec{\phi}_{Mj})}{d_{Mv}^2} \, d_{Tv}^2 + \vec{\beta}_{Tv}^T \vec{\phi}_{Tj}$ 
    \hfill Eq.~(21)
}
\hspace{0.3cm}

\KwRet{$\{ \hat{y}_{M1v}, \hat{y}_{M2v}, \dots, \hat{y}_{MJ_Mv} \}$}
\end{algorithm}

\vspace{1cm}

\begin{algorithm}[H]
\caption{Harmonization-Qc (with Bhattacharya Distance)}
\KwIn{}
    $\hat{D}_{Mv} = \{ (\hat{y}_{M1v}, \vec{x}_{M1}), \dots, (\hat{y}_{MN_Mv}, \vec{x}_{MJ_M'}) \}$ \tcp*{\small Harmonized data} 
    $D_{Tv} = \{ (y_{T1v}, \vec{x}_{T1}), \dots, (y_{TN_Tv}, \vec{x}_{TJ_T}) \}$ \tcp*{\small Target site data} 
    $thr$  \tcp*{\small Threshold} 
    $\vec{\beta}_{Tv}$ \tcp*{\small Target site parameter}
\KwOut{}
$d_{B}$ \tcp*{\small Bhattacharya Distance}

\hrulefill

\textbf{Data rectification:} \\
$Z_T = \{ z_{Tjv} = y_{Tjv} - \vec{\phi}_{Tj}^T \vec{\beta}_{Tv} \;|\;\;\; \forall (y_{Tjv}, \vec{\phi}_{Tj}^T) \in D_{Tv} \}$ \\
$Z_M = \{ \hat{z}_{Mjv} = \hat{y}_{Mjv} - \vec{\phi}_{Mj}^T \vec{\beta}_{Tv} \;|\;\;\; \forall (\hat{y}_{Mjv}, \vec{\phi}_{Mj}^T) \in \hat{D}_{Mv} \}$ \\

\vspace{0.3cm}
\textbf{Compute the Bhattacharyya distance:} \\
$\mu_T = \mathrm{mean}(Z_T), \quad \sigma_T^2 = \mathrm{var}(Z_T)$ \\
$\mu_M = \mathrm{mean}(Z_M), \quad \sigma_M^2 = \mathrm{var}(Z_M)$ \\
$d_B = \frac{1}{4} \frac{(\mu_T - \mu_M)^2}{\sigma_T^2 + \sigma_M^2} + \frac{1}{2} \ln \left( \frac{\sigma_T^2 + \sigma_M^2}{2 \sigma_T \sigma_M} \right)$ \\

\vspace{0.3cm}

\KwRet{$d_{B}$}
\end{algorithm}

\vspace{1cm}

\begin{algorithm}[H]
\caption{Hyperparameter Auto-Tuning}
\KwIn{}
    $D_{Tv} = \{ (y_{T1v}, \vec{x}_{T1}), \dots, (y_{TN_Tv}, \vec{x}_{TJ_T}) \}$ \tcp*{\small Target site data}
    $D_{Mv} = \{ (y_{M1v}, \vec{x}_{M1}), \dots, (y_{MN_Mv}, \vec{x}_{MJ_M}) \}$ \tcp*{\small Moving site data}
    $P, \nu_0, \tau_1, \tau_2, k, \lambda_{\min}$ \tcp*{\small Hyperparameters}

\KwOut{}
$\vec{\lambda}$ \tcp*{\small Harmonization parameters}

\hrulefill

\textbf{Initialization of $\vec{\lambda}$:} \\
$\{ d_{Mv}^2, \vec{\beta}_{Mv}, d_{Tv}^2, \vec{\beta}_{Tv} \} \leftarrow \text{ClinicalComBAT-Fit}(P, \vec{0}, \nu_0, D_{Tv}, D_{Mv})$ \\
$\vec{\lambda}_0 \leftarrow \left| \frac{\vec{\beta}_{Tv}^{[0]}}{\vec{\beta}_{Tv}} \right|$ \\

$found \leftarrow False$ \\
$\lambda \leftarrow \lambda_{\min}$ \\
\vspace{0.3cm}
\While{$found = False$}{
    $\vec{\lambda} \leftarrow \lambda \cdot \vec{\lambda}_0$ \\
    $\{ d_{Mv}^2, \vec{\beta}_{Mv}, d_{Tv}^2, \vec{\beta}_{Tv} \} \leftarrow \text{ClinicalComBAT-Fit}(P, \vec{\lambda}, \nu_0, D_{Tv}, D_{Mv})$ \\
    Compute $d_{\min}, d_{\max}, d_1, d_2$ \\
    
    \If{$ \text{sign}(d_{\min}\tau_1 - d_1) + \text{sign}(d_2 - d_{\max}\tau_2) + 2 = 0$ \hspace{1cm}Eq.~(35)} {
        $found \leftarrow True$
    }
    \Else{
        $\lambda \leftarrow \lambda \times k$ \tcp*{$k > 1$ is the multiplicative step size}
    }
}
\vspace{0.3cm}

\KwRet{$\vec{\lambda}$}
\end{algorithm}

\section{Supplementary Figures}

\begin{figure}[t]
\centering
\includegraphics[width=0.8\textwidth]{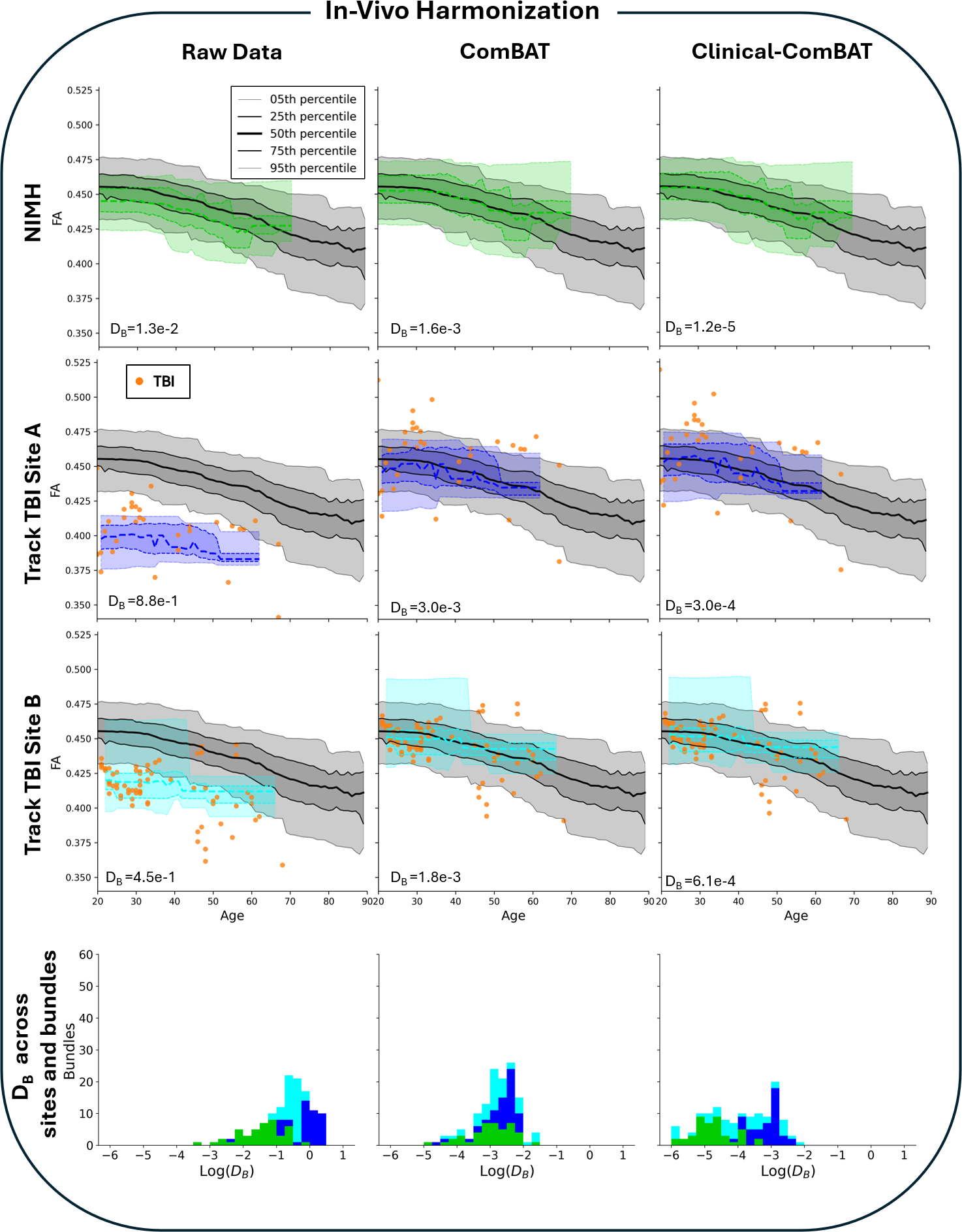}
\caption{Harmonization of fractional anisotropy (FA) using Clinical-ComBAT and ComBAT for the NIMH site (green), and Track-TBI site A (blue) and site B (cyan). The top three rows show harmonized white matter skeleton masks with corresponding $D_B$ values, where orange dots indicate TBI subjects. The bottom row presents stacked histograms of $D_B$ across all white matter bundles and sites. Lower $D_B$ values indicate closer alignment with the reference site, underscoring both the necessity of harmonization and the superior performance of Clinical-ComBAT over ComBAT.}\label{sfig:results-fa}
\end{figure}

\begin{figure}[t]
\centering
\includegraphics[width=0.8\textwidth]{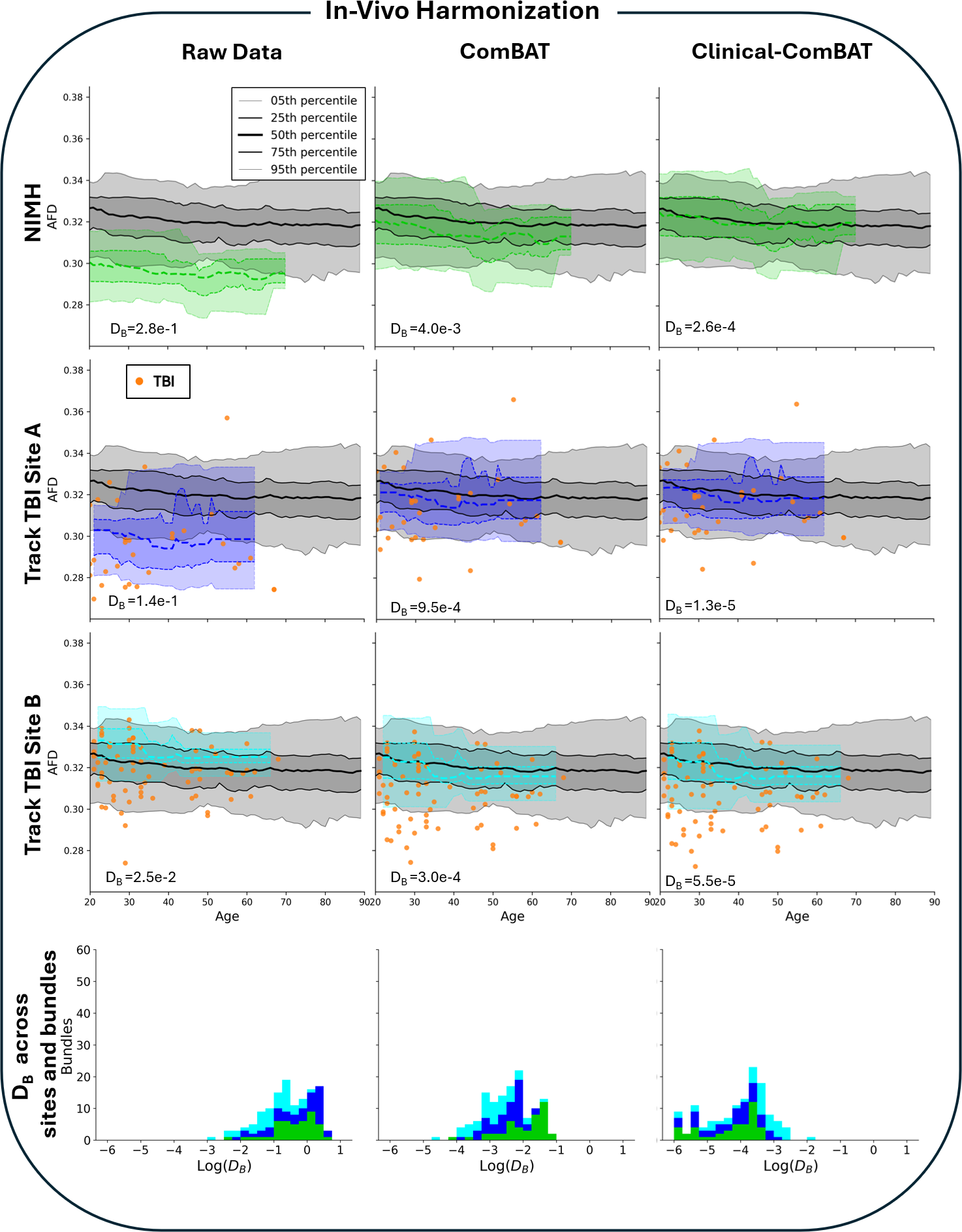}
\caption{Harmonization of Apparent Fiber Density (AFD) using Clinical-ComBAT and ComBAT for the NIMH site (green), and Track-TBI site A (blue) and site B (cyan). The top three rows show harmonized white matter skeleton masks with corresponding $D_B$ values, where orange dots indicate TBI subjects. The bottom row presents stacked histograms of $D_B$ across all white matter bundles and sites. Lower $D_B$ values indicate closer alignment with the reference site, underscoring both the necessity of harmonization and the superior performance of Clinical-ComBAT over ComBAT.}\label{sfig:results-afd}
\end{figure}


\end{document}